%% file: main.tex
\documentclass[10pt,journal, onecolumn, twoside]{IEEEtran}
\usepackage{times}
\usepackage{tabularx}
\usepackage{subcaption}
\usepackage{multirow,multicol, array}
\usepackage[font={small}]{caption}   
\usepackage{graphicx,dblfloatfix}
\usepackage{wrapfig}
\usepackage{siunitx}
\usepackage{lineno}

\usepackage{enumitem}

\usepackage{amsmath,amsthm,amssymb,amsfonts}
\usepackage[titlenumbered,ruled]{algorithm2e}
\usepackage{textcomp}
\usepackage{acronym}
\usepackage{balance}
\usepackage{mdwmath}
\usepackage{bm}
\usepackage{bbm}
\usepackage{mdwtab}
\usepackage{array}
\usepackage{eqparbox}
\usepackage{cite}

\usepackage{color}
\usepackage{psfrag}
\usepackage{epsfig}
\usepackage{url}

\usepackage{epstopdf}
\usepackage{booktabs}
\usepackage{blindtext}
\usepackage[colorlinks,bookmarksopen,bookmarksnumbered,linkcolor=blue,citecolor=blue,urlcolor=blue]{hyperref}

\usepackage[a-1b]{pdfx}

\renewcommand{\emph}{\textit}

\newtheorem*{lemma*}{Lemma}

\newtheorem*{problem*}{Problem}

\newcommand{\add}[1]{\textcolor{black}{#1}}

\newcommand{\modd}[1]{\textcolor{black}{#1}}

\allowdisplaybreaks[4]

\makeatletter
\newcommand\fs@spaceruled{\def\@fs@cfont{\bfseries}\let\@fs@capt\floatc@ruled
    \def\@fs@pre{\vspace{5\baselineskip}\hrule height.8pt depth0pt \kern2pt}%
    \def\@fs@post{\kern2pt\hrule\relax}%
    \def\@fs@mid{\kern2pt\hrule\kern2pt}%
    \let\@fs@iftopcapt\iftrue}
\makeatother

\IEEEoverridecommandlockouts

\begin{document}

%
\title{Robot-assisted Soil Apparent Electrical Conductivity Measurements in Orchards}
%
%
%

\author{Dimitrios Chatziparaschis,$^1$ Elia Scudiero,$^2$ and Konstantinos Karydis$^1$
\thanks{$^{1}$~D. Chatziparaschis and K. Karydis are with Dept. of Electrical and Computer Engineering, Univ. of California, Riverside, 900 University Avenue, Riverside, CA 92521, USA.
{\tt\footnotesize\{dchat013, karydis\}@ucr.edu}.}%
\thanks{$^{2}$~E. Scudiero is with Dept. of Environmental Sciences, Univ. of California, Riverside, 900 University Avenue, Riverside, CA 92521, USA.
{\tt\footnotesize elia.scudiero@ucr.edu}. }
\thanks{
We gratefully acknowledge the support of USDA-NIFA grant \# 2021-67022-33453, ONR grant \# N00014-19-1-2252, a UC MRPI Award, a Frank G. and Janice B. Delfino Agricultural Technology Research Initiative Seed Award, and an OASIS-IFA (Opportunities to Advance Sustainability, Innovation, and Social Inclusion – Internal Funding Awards). 
Any opinions, findings, and conclusions or recommendations expressed in this material are those of the authors and do not necessarily reflect the views of the funding agencies.
}}

\markboth{}
{Chatziparaschis \MakeLowercase{\textit{et al.}}: Robot-assisted Soil Apparent Electrical Conductivity Measurements in Orchards}

\maketitle

\begin{abstract}
Soil apparent electrical conductivity (ECa) is a vital metric in \modd{Precision Agriculture and Smart Farming}, as it is used for optimal water content management, geological mapping, and yield prediction. Several existing methods seeking to estimate soil \modd{electrical conductivity} are available, including physical soil sampling, ground sensor installation and monitoring, and the use of sensors that can obtain proximal ECa estimates. However, such methods can be either very laborious and/or too costly for practical use over larger field canopies. Robot-assisted ECa measurements, in contrast, may offer a scalable and cost-effective solution. In this work, we present one such solution that involves a ground mobile robot equipped with a customized and adjustable platform to hold an Electromagnetic Induction (EMI) sensor to perform semi-autonomous and on-demand ECa measurements under various field conditions. The platform is designed to be easily re-configurable in terms of sensor placement; results from testing for traversability and robot-to-sensor interference across multiple case studies help establish appropriate tradeoffs for sensor placement. Further, a developed simulation software package enables rapid and accessible estimation of terrain traversability in relation to desired EMI sensor placement. Extensive experimental evaluation across different fields demonstrates that the obtained robot-assisted ECa measurements are of high linearity compared with the ground truth (data collected manually by a handheld EMI sensor) by scoring more than $90\%$ in Pearson correlation coefficient in both plot measurements and estimated soil apparent electrical conductivity maps generated by kriging interpolation. The proposed robotic solution supports autonomous behavior development in the field since it utilizes the \add{Robot Operating System} (ROS) navigation stack along with the Real-Time Kinematic (RTK) GNSS positioning data and features various ranging sensors.
\end{abstract}

\IEEEpeerreviewmaketitle

\input{introduction.tex}

\input{background.tex}

\input{experiments.tex}

\input{fieldtests.tex}

\input{conclusion.tex}


\ifCLASSOPTIONcaptionsoff
  \newpage
\fi

\bibliographystyle{ieeetr} 

\end{document}

%% file: introduction.tex
\section{Introduction}

%
%
%
%

Agricultural geophysics employs non-invasive sensing techniques to characterize soil spatial variability and provide valuable insights into soil-plant-management relationships~\cite{geologyofsoils, birkeland1984soils}. Specifically, geospatial information on soil characteristics can indicate optimal cultivation approaches and may provide an estimate of expected yields~\cite{7,Tanji2002}. Soil salinity (i.e. salt content) is a crucial metric used to describe the soil characteristics and water content of an area. As such, it has been used widely across applications such as in agriculture, water management, geological mapping, and engineering surveys~\cite{8,huang2017time,soilproperties}. Information about bulk density, minerals content, pH, soil temperature, \modd{and more,} can be evaluated by measuring the \modd{ECa} of the field and generating a profile of the surveyed land. Thus, approximating the \modd{ECa} spatial variability of a field can provide a broader understanding of the water flow through the ground, pinpoint any spots with irregular \modd{soil patterns}, and finally indicate the necessity of supplying additive plant nutrients or different irrigation approaches~\cite{https://doi.org/10.2134/agronj2003.4550}.

\begin{figure*} [t!]
    \centering
  \subfloat[\label{fig:handheld}]{%
       \includegraphics[width=0.3\linewidth]{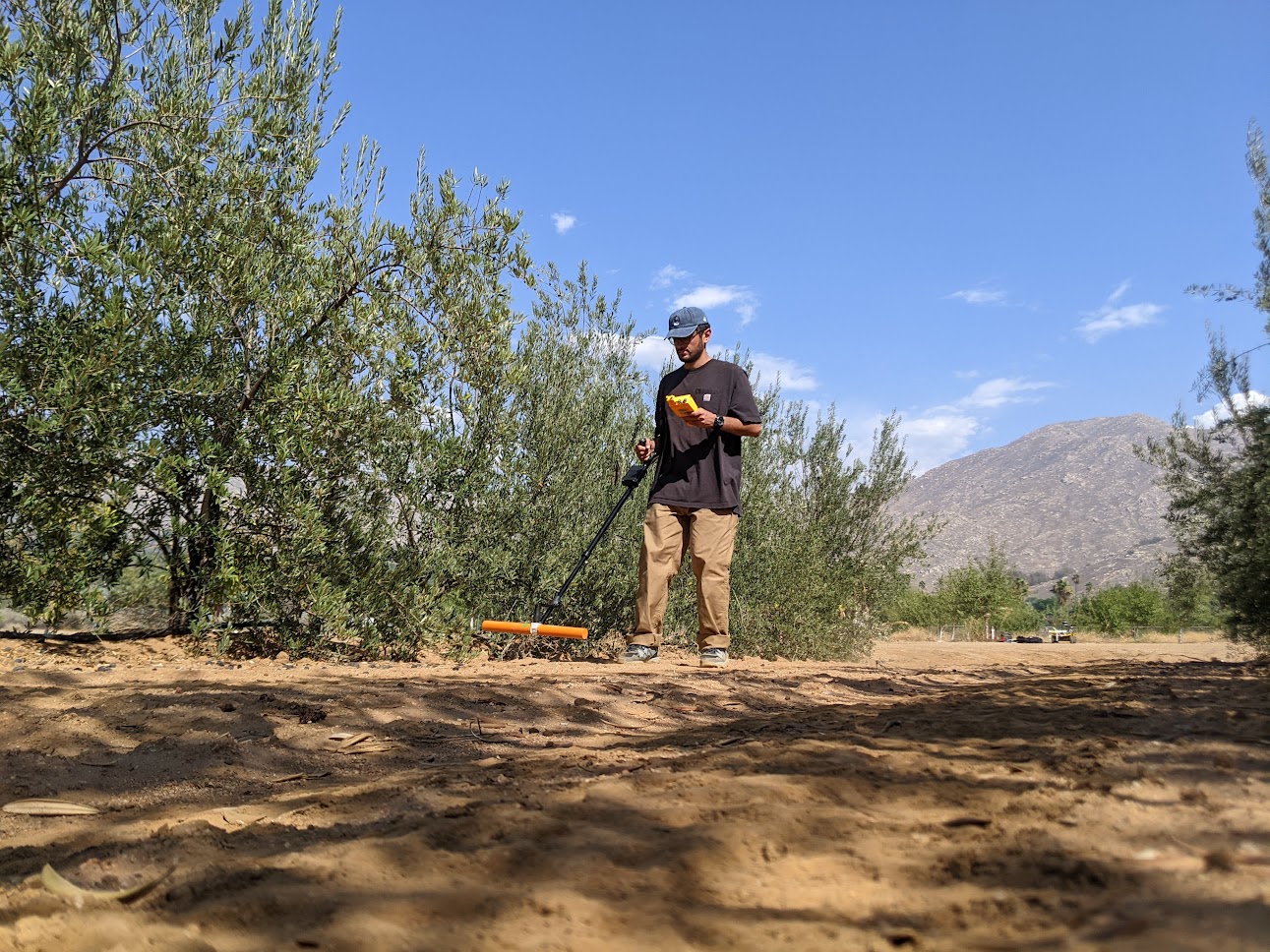}}
    \hfill
      \subfloat[\label{fig:jackalandemi}]{%
    \includegraphics[width=0.3\linewidth]{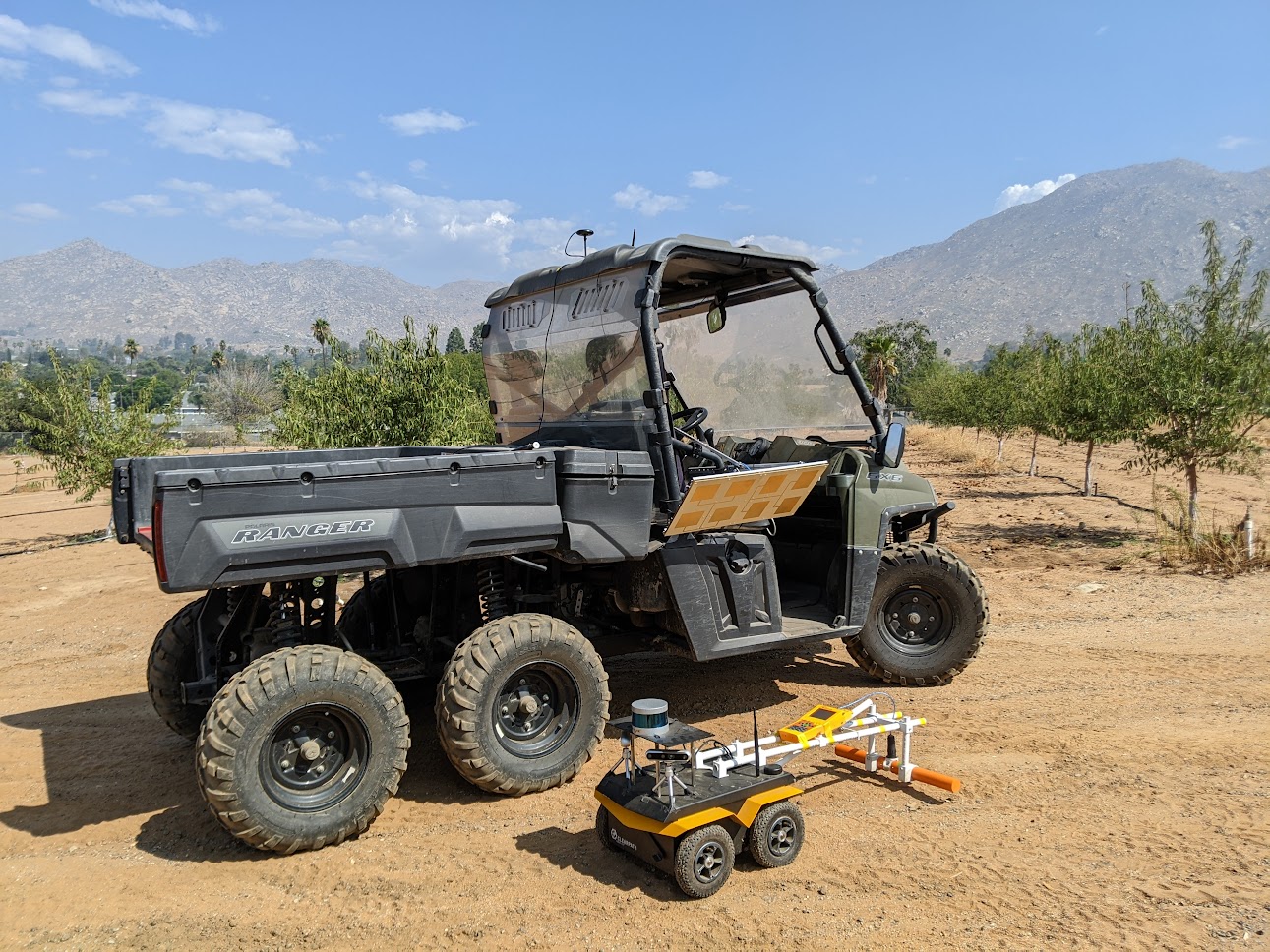}}
        \hfill
      \subfloat[\label{fig:rtkatsalinity}]{%
    \includegraphics[width=0.3\linewidth]{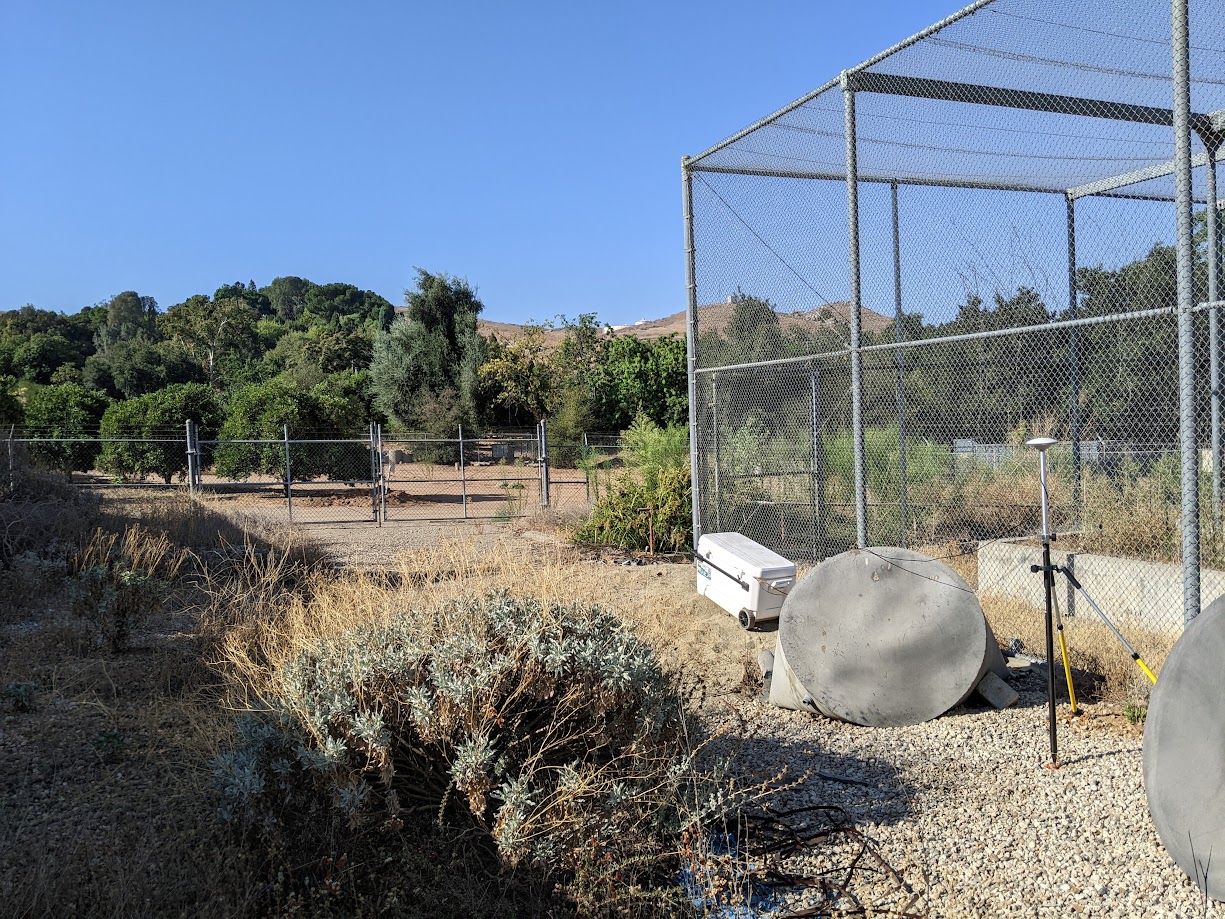}}
  \caption{(a) Instance of the manual data collection process using the handheld EMI sensor. Manually-collected data serve as ground truth in this work. (b) The robot considered in this work is a Clearpath Jackal UGV wheeled robot, carrying a GF CMD-Tiny EMI instrument (long orange cylinder) for ECa measurements \modd{alongside a Polaris ATV}. (c) GNSS positioning information for the robot to navigate autonomously as well for geo-localization and cross-reference of obtained sensor measurements is provided via an RTK-Base station. \label{fig:overview}} 
\end{figure*}

In general, \modd{soil conductivity} is mainly estimated in-situ, using three distinctive methods~\cite{adamchuk2018proximal}\add{;} moisture meters (hydrometers) installed into the ground, time-domain reflectometers, and measurement of soil electromagnetic induction (EMI). In the first two cases, growers install and use decentralized sensor arrays to gather information from selected points on the field \cite{reflectometry,sarray1,sarray2}. A main drawback of these approaches is that they provide discrete measurements and hence sparse information over the complete field, which may lead to less efficient  agricultural tactics and higher costs while aiming to scale over larger fields. On the other side, ECa is measured geospatially (on-the-go) with Electrical Resistivity methods and EMI sensors. The EMI measurements of soil apparent electrical conductivity can be performed in a continuous manner and proximally, whereby a farm worker walks through the field holding the EMI sensor or a field vehicle that carries the sensor is driven around~\cite{8} (Fig.~\ref{fig:overview}). In this way, a more spatially-dense belief about field irrigation is formed as the sensor can either gather continuous measurements of selected regions within the field or sparse measurements from specific points. Figure~\ref{fig:handheld} depicts an instance from a manual survey of soil moisture in \modd{an} olive tree field, with the use of the GF CMD-Tiny EMI instrument. Despite the benefits afforded by continuous EMI soil measurements, a noteworthy drawback is that such surveys may not scale well in larger fields as they can become quite labor-intensive depending on the broader area of inspection and weather conditions (e.g., heat fatigue). \modd{The standard of practice (besides manual operation) is to use an ATV (Fig.~\ref{fig:jackalandemi}) that carries a (often larger) sensor; however, due to its size, it may be hard to get the sensor close to the tree roots where estimating soil apparent electrical conductivity is most crucial. A robot-assisted solution has the potential to get the sensor close to the roots (as in the manual case) in a less laborious way (as in the ATV case) and thus bridge the gap between the two main standard of practice methods by offering an attractive alternative.}

The use of (mobile) robots has been offering key assistance in contemporary survey and agronomy processes, for example to better understand field conditions with the use of onboard sensors, provide real-time field modeling and decision-making on agricultural tactics, and at cases offer aid to farm workers (e.g., transporting workers across the field or elevating them to reach parts of the tree that are high off the ground~\cite{https://doi.org/10.1002/rob.22106}). In many scenarios, Unmanned Ground Vehicles (UGVs) are used to collect data from the ground~\cite{9823443, 10.1117/12.2618728,9277577}, while other studies utilize aerial data taken by Unmanned Aerial Vehicles (UAVs)~\cite{8909233, doi:10.1142/S2301385022500029, 8719237} and make decisions\cite{7535938, 9791350} for the surveying area on-the-fly. As a follow-up, there is research that demonstrates collaborative approaches with both aerial and ground robots~\cite{7587351,10.1117/12.2622615,8620543} which can yield an even more detailed and broader inspection of the field. Out of all types of data collected in the field, the most relevant to this present work concerns soil moisture.


During the past years, there have been various approaches aiming to utilize robots for collection of soil moisture measurements. Some examples of related research include mobile robots conducting soil potential of hydrogen (pH) measurements for determination of soil health in the survey site~\cite{s110100573}, robot manipulators with onboard depth cameras that place soil moisture sensors on the plants~\cite{DBLP:journals/corr/abs-2201-07653}, and even a UAV-based multispectral system for estimating and modeling soil salinity of the inspected field~\cite{s22020546}. Thayer $et~al.$~\cite{8461242} revealed the NP-hard routing problem of autonomous robots in precision irrigation scenarios and developed two domain-specific heuristic approaches.

%
%
In more detail, Pulido Fentanes $et~al.$~\cite{12} demonstrated a mobile robot that performs soil moisture measurements autonomously in the field, using a cosmic-ray sensor. The main drawback of using such sensory equipment is the increase in the overall cost of the application, which might be accurate but financially inefficient for some farming applications. 
On the aerial side, Tseng $et~al.$~\cite{tseng2018towards} showed the usability of machine learning with aerial imagery to learn and predict the soil moisture conditions on individual plants and large fields through the air. Even though this application reported reduced water consumption by up to 52\%, there is a need for a more accurate and robust system for measuring soil moisture levels.
In addition, Lukowska $et~al.$~\cite{8765937} presented an all-terrain six-wheeled robot, that utilizes a drilling system with a flap design to perform soil sampling in larger fields. Along similar lines, Dimaya $et~al.$~\cite{8666240} developed a mobile soil robot collector (SoilBot) to automate soil collection in a sugar cane field. In such cases, as the post-processing of the soil samples will provide detailed information about the moisture level of the sampled soil, there may be approaches for real-time and broader field monitoring robotic applications. 
%
In addition, $Agrobot$~\cite{9823443} is a farm robot that has been designed to reduce human labor through autonomous seed sowing and soil moisture measuring. Further, Bourgeois $et~al.$~\cite{9668355} demonstrated a low-cost and portable land robot, namely RoSS, for soil quality sensing purposes, which can collect soil samples and/or insert soil sensor probes at sampling locations to measure the moisture levels. 
In these approaches, vital information for the total salinity of the field might be missing as local information is obtained from the water content in the soil of the sampled areas. Campbell $et~al.$~\cite{campbell} showcased a small-factor robot with an attached EMI sensor for field-scale soil apparent electrical conductivity measurements. Even though this work presents an efficient approach toward integrating an EMI sensor onto a mobile robot to conduct continuous ECa measurements, it has limited traversability and operational time because of its small size.

In this study we present a mid-sized ground mobile robot solution that is able to conduct semi-autonomous and on-demand continuous EMI ECa measurements under various and larger field environments, and obtain a field-scale ECa map. Our proposed solution is based on the Clearpath Jackal UGV, which is equipped with a customized and adjustable platform which can carry the EMI instrument GF CMD-Tiny. The robot supports teleoperation via Bluetooth, it can directly navigate through sending desired waypoints and can execute trajectories as it utilizes its onboard GPS and RTK positioning data along with local and global planners to reach desired goals. The design, hardware and software system integration and testing of this platform \modd{are} fully presented and evaluated in both simulated and real field-scale scenarios, including over bare fields and in muddy terrains. The proposed robot platform demonstrates high efficiency in terms of portability, traversability as well as data collection since the robot is found capable to collect data with high linearity compared to handheld (no-robot) cases. Thus, our proposed solution shows promise to serve as a useful tool in modern field survey, ECa mapping, and irrigation scheduling. 

In the remainder of the manuscript, we first discuss key components (off-the-shelf as well as fabricated in-house) and the overall system design, and offer key system integration information (Section~\ref{sec:system}). An important contribution of this work is a thorough study of the tradeoffs regarding EMI sensor placement on the mobile robot that involves tools and processes we develop for both experimental testing and testing in simulation; this is presented in Section~\ref{sec:development}. Details and key findings from an initial testing phase to validate the preliminary efficacy of the overall system are also reported in that section. Full system evaluation and testing results across multiple trials in two distinctive fields are presented in Section~\ref{sec:evaluation}. Finally, Section~\ref{sec:conclusion} concludes this manuscript.

%% file: background.tex
\section{System Design and Integration of Key Components}\label{sec:system}

\subsection{Soil Conductivity and Employed Sensor}

Electromagnetic induction can help measure the soil apparent electrical conductivity of a field. The main operating principle is based on the evaluation of the induced magnetic field from the ground as transmitted by an electromagnetic conductivity meter. Specifically, the EMI transmitter emits a harmonic signal toward the ground and generates a magnetic field. Since the receiver is placed with the same dipole orientation as the transmitter, it captures the secondary (induced) magnetic field which relates to the ground conductivity, namely out-of-phase measured in $mS/m$ and the in-phase that is a relative metric to the primary magnetic field and measures the magnetic susceptibility of the area.

In our study, we use the CMD-Tiny meter from GF Instruments, which features a compact and lightweight build. This instrument has a control unit module that is used for configuring and logging the EMI measurements and the CMD probe that is the main magnetic sensing module. The latter component has a cylindrical shape with $50\;cm$ length and $4.25\;cm$ of diameter, and the total setup weighs $425\;g$ (Fig.~\ref{fig:handheld}). This EMI instrument can obtain
soil conductivity measurements from $0.35\;m$ up to $0.7\;m$ in-ground depth according to its setup and selected resolution.

\subsection{Mobile Robot Setup}
We use the Clearpath Robotics Jackal robot platform,\footnote{~\url{https://clearpathrobotics.com/jackal-small-unmanned-ground-vehicle/}} which is a UGV designed for use in outdoor and rugged all-terrain environments. The Jackal has been used in agricultural robotics research~\cite{10.1117/12.2623034,LI2016341}, autonomous exploration~\cite{9810192, 9830880,9829293}, as well as social-aware navigation~\cite{8202312,9340710}. 
The robot's dimensions are $50.8\times 43.2\times 25.4~cm$ with a payload area of $43\times 32.25~cm$ for mounting various (OEM and custom) onboard modules. Its available payload capacity reaches $20\;kg$. The robot features an onboard NVIDIA Jetson AGX Xavier computer that is responsible for all onboard computation. On the sensors and actuators side, the robot is equipped with a GNSS receiver, an IMU module, and motorized wheel encoders (besides the custom payloads developed in this work and which we discuss later \add{, or additional sensors like stereo cameras and LiDAR that are routinely deployed on the robot for autonomous navigation~\cite{karnan2022voila,wang2021apple}).} Additionally, the Jackal is a ROS-compatible robot, as it uses the ROS navigation stack and the ROS environment for its main functionality. The total operating time of this robot can reach up to $4\;hrs$ depending on the use and type of the operating environment. \add{Importantly, the required operating time on the field can vary depending on field size and type, and the desired field mapping resolution. For instance, the surveys conducted herein took place over a total period of $1.5\;hrs$ (3 surveys each lasting for $0.5\;hrs$), in a $30\times15\;m$ field, and with the robot moving with a stable linear speed of $1\;m/s$. Under these settings, the total battery consumption never exceeded $40\%$ during our experiments in each case, for performing the total ECa mapping of the field. The operational time of the Jackal can be expanded by the use of a secondary/additional battery. If an increased amount of surveying time and operation in even larger fields are desired, an alternative commercial wheeled robot can be employed instead (e.g., the Clearpath Robotics Husky,\footnote{~\url{https://clearpathrobotics.com/husky-unmanned-ground-vehicle-robot/}} and Warthog UGV,\footnote{~\url{https://clearpathrobotics.com/warthog-unmanned-ground-vehicle-robot/}} or the Amiga from Farm-ng\footnote{~\url{https://farm-ng.com/products/la-maquina-amiga}}); our method can directly apply to such robots as well.} 
Figure~\ref{fig:jackalandemi} depicts the Jackal robot equipped with various onboard sensors, in comparison to a Polaris ATV that is typically used for mechanized EMI measurements.

\subsection{Positioning System Integration for Field Navigation}

Nowadays, high-in-accuracy Global Navigation Satellite Systems (GNSS) are used (along with field sensors) in precision agriculture to generate, extract, and obtain field observations with spatial information. In many cases, Real-Time Kinematic (RTK) positioning is applied to provide $cm$-level accuracy on the captured data by using real-time corrections through an established and calibrated base station.

Herein we use the Holybro H-RTK F9P GNSS series as the high-in-accuracy positioning module for our field experiments, which
integrates a differential high-precision multi-band GNSS positioning system. By selecting specific points on the field, we calibrate the RTK base station to obtain its position at $cm$-level and we use telemetry to establish communication with the robot's onboard autopilot hardware. \add{Figure~\ref{fig:rtkatsalinity} illustrates an H-RTK F9P RTK base station establishment in an outdoor environment.} On the robot side, we use the Holybro Pixhawk 4 autopilot module in the rover airframe (UGV mode), along with the Holybro SiK Telemetry V3 $100\;Mhz$ and the GNSS receiver to capture the RTK corrections from the base station. 
The MAVROS software\footnote{~\url{http://wiki.ros.org/mavros}} is utilized to parse the positioning data and use them with the onboard computer through a USB connection at a rate of $10\;Hz$. In this way, the Jackal robot is able to reliably georeference every captured measurement in the field.

On the navigation side, the robot's captured positions are described in the World Geodetic System 1984 (WGS-84). Additionally, the Jackal uses an extended Kalman Filter\cite{ekf}, which fuses information from the onboard IMU and the wheel encoders to provide the state estimation of the robot (odometry). Since in our case we require the Jackal to be able to follow a predefined GNSS-tagged trajectory, we utilize the $navsat\_transform\_node$\footnote{\label{foot:navsat}~\url{http://docs.ros.org/en/jade/api/robot\_localization/html/navsat\_transform\_node.html}} from the ROS navigation stack. Through this approach, initially, all the requested geotagged targets are transformed to the Universal Transverse Mercator (UTM) coordinate system.  Given this information and through the positioning data of the robot's odometry, a static transformation is generated to describe both the UTM coordinates of the robot and the targets' position into the local robot's frame. In this way, the requested trajectory can be georeferenced and transformed into Jackal's local frame and thus the robot can follow it. 
\add{It is worth mentioning that, in case the robot gets into a position where there is limited satellite visibility (i.e. GPS-denied environments), it continues geotagging the captured measurements based on pose belief estimation via fusion of the onboard IMU and wheel encoders odometry data.\footnotemark[6] While not employed in this work, our method is autonomous-ready in the sense that it can directly be integrated with waypoint navigation determined readily by onboard sensors (e.g.,~\cite{mohta2018fast}), where task allocation and motion planning for (newly-perceived) obstacle avoidance can happen online (e.g.,~\cite{kan2020online,kan2021task}).} \add{Also, mapped modalities from the flora~\cite{campbell2022integrated} and the fauna~\cite{ye2020development} during the survey, can even enhance and provide a multi-modal belief about the field conditions.}

\add{Additionally, on the traversing side, we focus on open-world navigation without the existence of obstructive objects in the path that may cause the robot to get out of track to bypass them. The robot can get over small tree branches or crops as a farm worker would do, but the ground should be relatively uncluttered for the ECa inspection as in a typical survey scenario. Also, as Jackal is an IP62 rugged robot, it has a capable high torque  $4\times4$ drivetrain allowing it to navigate through muddy and uneven parts similar to a common ATV. In our experiments, we test our proposed system in both dry and muddy environments to demonstrate our system's efficacy. The robot supports a $20\;cm$ wheel diameter (up to  $30\;cm$) and a variety of tire models, such as square (plastic) spike sets, which can make it even more versatile for challenging terrain types that require higher traction.}

\subsection{Robot Configuration and the Design of the Sensor Mounting Platform}

\begin{figure}[t!] 
    \centering
  \subfloat{%
       \includegraphics[width=0.33\linewidth]{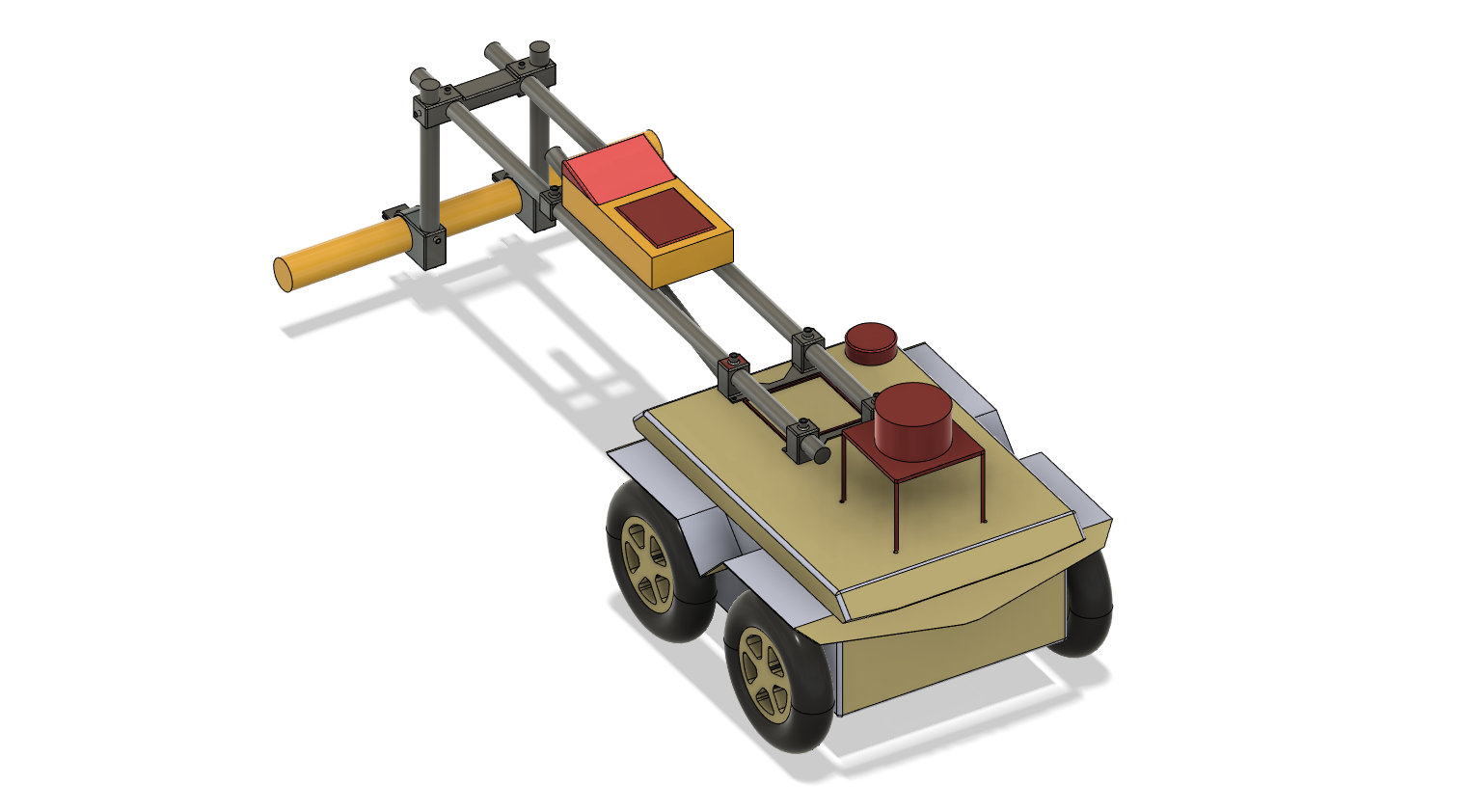}}
    \hfill    
  \subfloat{%
       \includegraphics[width=0.33\linewidth]{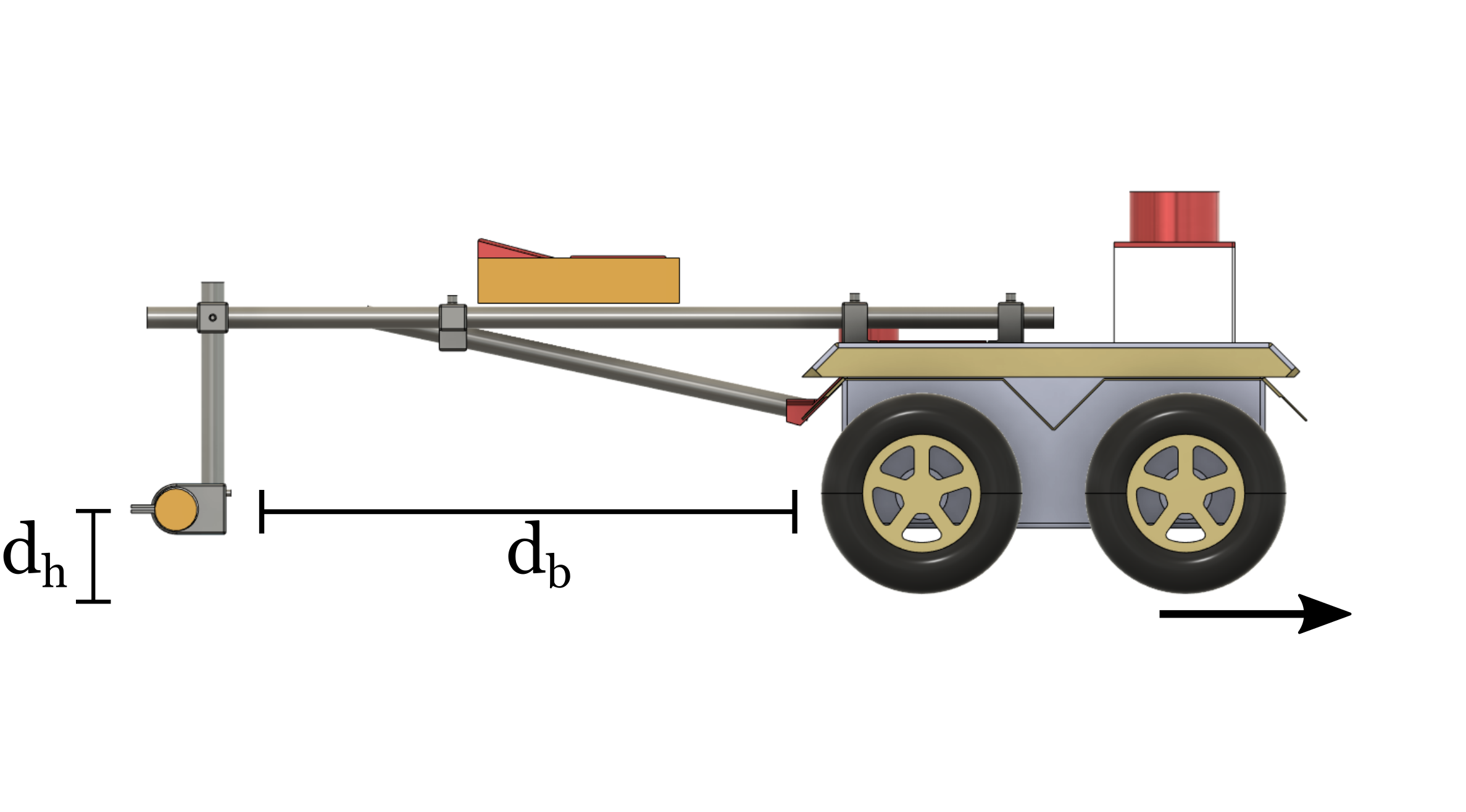}}
    \hfill
  \subfloat{%
       \includegraphics[width=0.33\linewidth]{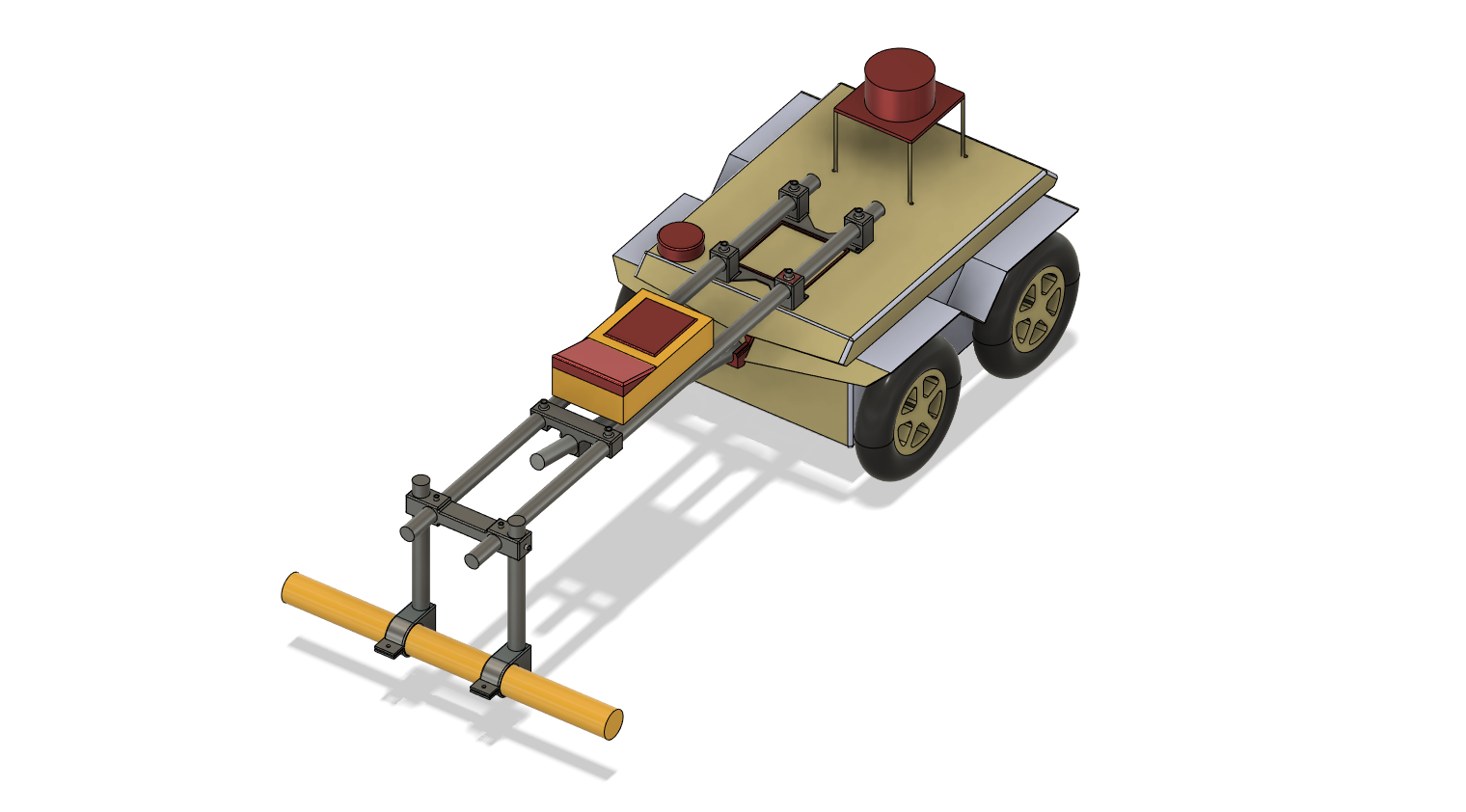}}
  \caption{CAD rendering of the Jackal's platform to hold the GF CMD-Tiny instrument. The parameters $d_h$ and $d_b$ indicate the adjustable height and distance of the sensor probe, respectively.\vspace{-0.5cm} \label{fig:fusion}} 
\end{figure}

As our aim is to attach the CMD-Tiny meter on the Jackal robot, we start with the design of the sensor platform and the definition of its adjustable parameters. The prototype renderings are depicted in Fig.~\ref{fig:fusion}. As the Jackal robot has a limited payload area, we aim to design and use a module that can be attached on the robot's top plate and extend outwards so that the sensor can get in close proximity to the ground. In this way, the robot can carry the EMI sensor and obtain soil conductivity measurements while navigating in a continuous manner. However, the sensor's sensitivity in magnetic field measurements to other metallic and/or electronic components (like the robot itself) in view of module placement on the robot requires a study of some key tradeoffs.

In particular, the EMI measurements can be distorted by the presence of other magnetic fields caused by the robot, as well as by random oscillations caused by the robot's movement. Thus, one intuitive solution would be to place the sensor as far as possible from the robot (i.e. large value for parameter $d_b$) and as close to the ground as possible (i.e. small value for parameter $d_h$). However, as distance $d_b$ increases (Fig.~\ref{fig:fusion}), the moment arm of the payload increases, which in turn can lead to higher power consumption of the robot. In addition, the magnitude of the oscillations (which directly affects EMI measurement consistency) will also increase. Finally, the longer the distance the worse the capability of the robot to traverse uneven terrain and/or negotiate dips and bumps that are bound to exist in the field in practice, as possible sensor collisions with the ground can occur if height $d_h$ is not sufficiently large. For these reasons, we design an adjustable platform to support multiple configurations of the CMD-Tiny meter's control unit and probe position, relative to the robot's main body, and study the tradeoff between measurement consistency and robot traversability to identify an optimal sensor placement (see Section~\ref{sec:development} below).

Figure~\ref{fig:fusion} provides multiple views of the designed platform in CAD (Fusion 360), with all the designed parts included in the final assembly. PVC tubes of $80\;cm$ length, $10\;mm$ diameter, and $SCH~80$ wall thickness serve as the basis for an expandable and rigid structure to mount the sensor (cylindrical part) as well as its associated data collection logger (rectangular part placed on top of the two long PVC tubes). Additionally, a shorter PVC tube of $30\;cm$ length, has been installed along with an intermediate support mounted on the Jackal's front bumper to enhance the stability of the overall EMI module. On the robot side, two supports are installed on the Jackal's top chassis to mount the two PCV tubes, and a unified support has been installed on the opposite side to hold the sensor's cylindrical probe. The latter support is crucial in allowing for reconfiguring the position and height from the ground of the probe and can be sturdily fixed in place when in use. Also, the total length of the sensor holder platform can be modified by the supports that are installed on the top chassis. Fabricated components of the platform were all 3D printed in polylactic acid (PLA) material.

%% file: experiments.tex
\section{Development and Calibration for Optimal Sensor Placement}\label{sec:development}

The optimal sensor placement in terms of the robot body is determined by two factors: 1) electromagnetic interference from the robot chassis and its onboard electronics, and 2) robot terrain traversability. The optimal solutions to each of these factors on their own are in fact opposing each other. Indeed, to minimize electromagnetic interference, the sensor should be placed as far from the robot as possible. On the contrary, the longer the extension of the sensor holding platform from the robot's center of mass, the worse it becomes to overcome obstacles without risking colliding the sensor with the ground (Fig.~\ref{fig:fusion}), in addition to increasing the moment arm of the cantilever which in turn would increase the required motor torque to move without tilting forward (that also leads to higher power consumption and lower operational time). In this section, we study optimal sensor placement in two distinct settings. First (Section~\ref{seq:sensorinterference}), we benchmark experimentally the robot's interference on the EMI measurements for a given fixed distance from the ground ($d_h=5\;cm$) and for varying distances of the probe ($d_b\in[10,100]\;cm$). 
Second (Section~\ref{seq:gazeboexperiments}), we determine terrain traversability while considering a subset of viable distances identified from the first step and two distinctive sensor height values to better understand the role of robot oscillations onto potential probe collisions with the ground. To make this process systematic, we create a realistic simulated environment and conduct this initial set of traversability evaluations in simulation. This process yields a set of candidate probe placement distances ($d_b$) as a function of probe height off the ground ($d_h$). Additionally, we perform feasibility experimental testing in measuring continuous ECa over small areas to both narrow down the range of viable configurations and to validate optimal sensor placement (Section~\ref{sec:validation}).

\begin{table*}[!t]
\caption{Soil Conductivity Measurements in Evaluating Robot Electromagnetic Interference. \label{tab:prwthmera}}
\centering
\begin{tabular}{cclcccccccccc}
\toprule
Field Type    & \textbf{}             & \textbf{} &
\multicolumn{10}{c}{Distance between the robot and the probe ($d_b$) for constant probe height off the ground ($d_h=5\;cm$)} \\ \midrule
\multicolumn{1}{l}{} & \multicolumn{1}{l}{} &           & \multicolumn{1}{l}{} & \multicolumn{1}{l}{} & \multicolumn{1}{l}{} & \multicolumn{1}{l}{} & \multicolumn{1}{l}{} & \multicolumn{1}{l}{} & \multicolumn{1}{l}{} & \multicolumn{1}{l}{} & \multicolumn{1}{l}{} & \multicolumn{1}{l}{} \\
$mS/m$                 & $\infty$                &           & $10\;cm$               & $20\;cm$               & $30\;cm$               & $40\;cm$               & $50\;cm$               & $60\;cm$               & $70\;cm$               & $80\;cm$               & $90\;cm$               & $100\;cm$              \\ 
\midrule
                       &                       &           &                      &                      &                      &                      &                      &                      &                      &                      &                      &                      \\
\multicolumn{1}{l}{}  & 13.9                  &           & 39.1                 & 85                   & 38                   & 26.2                 & 19.8                 & 17.1                 & 15.6                 & 14.9                 & 14.6                 & 14.6                 \\
                       & 9.9                   &           & 36.8                 & 58.4                 & 28.8                 & 18                   & 13.4                 & 11.6                 & 10.8                 & 10.4                 & 10.2                 & 10.1                 \\
\textit{Bare Field}    & 10.9                  &           & -2                   & 88.7                 & 38.8                 & 24.4                 & 16.2                 & 13.2                 & 12.1                 & 11.8                 & 11.4                 & 11.2                 \\
                       & 13.2                  &           & -17.8                & 78.4                 & 33.8                 & 22.5                 & 18.1                 & 15                   & 14.1                 & 13.7                 & 13.3                 & 13.1                 \\
\multicolumn{1}{l}{}  & 10.3                  &           & 66                   & 61.8                 & 30.3                 & 20.2                 & 15.4                 & 12.8                 & 11.5                 & 11                   & 10.7                 & 10.5                 \\
\midrule
\textit{Mean ($\mu$)}   &
11.64 & \multicolumn{1}{l}{} &24.42 & 74.46 & 33.94 & 22.26 & 16.58 & 13.94 & 12.82 & 12.36 & 12.04 & 11.90  \\
\textit{Standard Deviation ($\sigma$)}   &
1.80 & \multicolumn{1}{l}{} 
 & 33.83 & 13.67 & 4.47 & 3.26 & 2.46 & 2.15 & 1.98 & 1.89 & 1.85 & 1.90  \\
\midrule
\midrule
\multicolumn{1}{l}{}  & \multicolumn{1}{l}{} &           & \multicolumn{1}{l}{} & \multicolumn{1}{l}{} & \multicolumn{1}{l}{} & \multicolumn{1}{l}{} & \multicolumn{1}{l}{} & \multicolumn{1}{l}{} & \multicolumn{1}{l}{} & \multicolumn{1}{l}{} & \multicolumn{1}{l}{} & \multicolumn{1}{l}{} 
\\
                       & 25.3                  &           & 77.8                 & 90.2                 & 51.9                 & 37.4                 & 30.6                 & 28                   & 26.5                 & 26                   & 25.6                 & 25.5                 \\
                       & 26                    &           & 47.2                 & 96.1                 & 51.3                 & 36.3                 & 30.5                 & 28.1                 & 27                   & 26.6                 & 26.4                 & 26.2                 \\
\textit{Olive Tree Grove}  & 21.8                  &           & 7.8                  & 106.3                & 50.6                 & 34.1                 & 27.8                 & 24.4                 & 23.1                 & 22.4                 & 22.1                 & 21.9                 \\
                       & 30.3                  &           & 50.8                 & 105.3                & 55.2                 & 41.3                 & 35.2                 & 33.1                 & 31.6                 & 31                   & 30.8                 & 30.6                 \\
                       & 23.2                  &           & 59.4                 & 87.9                 & 48.1                 & 32.3                 & 29                   & 25.6                 & 24.5                 & 24.1                 & 23.8                 & 23.6 \\
\midrule
\textit{Mean ($\mu$)}   &
25.32 & \multicolumn{1}{l}{} 
 & 48.60 & 97.16 & 51.42 & 36.28 & 30.62 & 27.84 & 26.54 & 26.02 & 25.74 & 25.56 \\
\textit{Standard Deviation ($\sigma$)}   &
3.25 & \multicolumn{1}{l}{} 
 & 25.69 & 8.44 & 2.56 & 3.43 & 2.81 & 3.34 & 3.23 & 3.24 & 3.28 & 3.28  \\ 
\midrule 
\midrule
\multicolumn{1}{l}{}  & \multicolumn{1}{l}{} &           & \multicolumn{1}{l}{} & \multicolumn{1}{l}{} & \multicolumn{1}{l}{} & \multicolumn{1}{l}{} & \multicolumn{1}{l}{} & \multicolumn{1}{l}{} & \multicolumn{1}{l}{} & \multicolumn{1}{l}{} & \multicolumn{1}{l}{} & \multicolumn{1}{l}{}
\\
\multicolumn{1}{l}{}  & 27.3                  &           & 73.9                 & 90.7                 & 51.7                 & 35.7                 & 31.4                 & 29.2                 & 28.3                 & 27.9                 & 27.6                 & 27.5                 \\
                       & 23                    &           & 69.2                 & 85.2                 & 47.3                 & 33.6                 & 27.7                 & 25.3                 & 24.2                 & 23.6                 & 23.4                 & 23.2                 \\
\textit{Citrus Tree Grove} & 29.2                  &           & 73.1                 & 90.5                 & 53.5                 & 39                   & 33.6                 & 31.3                 & 30.2                 & 29.8                 & 29.5                 & 29.3                 \\
                       & 32.2                  &           & 35.6                 & 101.7                & 57.8                 & 43.1                 & 37.2                 & 34.7                 & 33.4                 & 32.8                 & 32.5                 & 32.3                 \\
                       & 22.7                  &           & 50.3                 & 96.5                 & 48.9                 & 33.1                 & 27.6                 & 25.3                 & 24.6                 & 23.4                 & 23.1                 & 23          \\ 
\midrule
\textit{Mean ($\mu$)}   &
26.88 & \multicolumn{1}{l}{}  & 60.42 & 92.92 & 51.84 & 36.90 & 31.50 & 29.16 & 28.14 & 27.50 & 27.22 & 27.06\\
\textit{Standard Deviation ($\sigma$)}   &
 4.07 & \multicolumn{1}{l}{} 
 & 16.87 & 6.33 & 4.11 & 4.17 & 4.08 & 4.03 & 3.87 & 4.05 & 4.02 & 4.00   \\
\bottomrule
\end{tabular}
\end{table*}

\subsection{Determination of Robot Interference in Soil Conductivity Measurements}\label{seq:sensorinterference}

The first step to be examined concerns the robot's electromagnetic interference in the soil conductivity measurements. Specifically, the robot is equipped with a high-torque $4\times4$ motored drivetrain and various onboard sensors such as a 3D LiDAR and the GNSS receiver. Given this setup, the robot generates electromagnetic fields that may interfere with the electromagnetic field generated by the GF CMD-Tiny sensor as well as the secondary current read by the sensor, and thus cause distorted measurements.

\subsubsection*{\textbf{Experimental Procedure}} To examine and mitigate the robot's interference on the EMI measurements, three independent field experiments were conducted under different robot-sensor configurations in which the sensor was \add{placed} at predefined positions away from the robot's body to monitor the level of saturation. The fields that were selected for this purpose included a bare field, an olive tree grove, and an orange tree grove, which are located at the USDA-ARS U.S. Salinity Laboratory at the University of California, Riverside (33$^\circ$58'21.936'' N, -117$^\circ$19'13.5732'' E). In each field experiment, five distinct field points were selected to measure the soil conductivity at. To get a broad spectrum of values for better understanding of the robot's electromagnetic interference, sampling points were either irrigated recently or non-irrigated. At the beginning of each experiment, handheld measurements were performed with the CMD-Tiny sensor, with no presence of any device that may generate additional electromagnetic fields. These measurements serve as the baseline. With these as a reference, we then placed the UGV with the CMD-Tiny sensor in different configurations and repeated the data collection process, each time increasing their relative distance within the range of $[10,100]\;cm$ at a step of $10\;cm$. 

Soil conductivity data collected from these field tests are shown in Table~\ref{tab:prwthmera}. Column ``$\infty$" represents the handheld measurements where there is no appearance of external electromagnetic interference; the reported values are used as reference measurements. The columns starting  from $10\;cm$ until $100\;cm$ represent the soil conductivity measurements that were made at the corresponding relative distances of the robot and the sensor's probe, at the same locations as in the handheld case. Mean and one-standard deviation values from the five trials in each of the 33 distinctive cases shown in Table~\ref{tab:prwthmera} are also provided.

\begin{figure*}[!t]
    \centering
  \subfloat[\label{s3-a}]{%
       \includegraphics[width=0.33\linewidth]{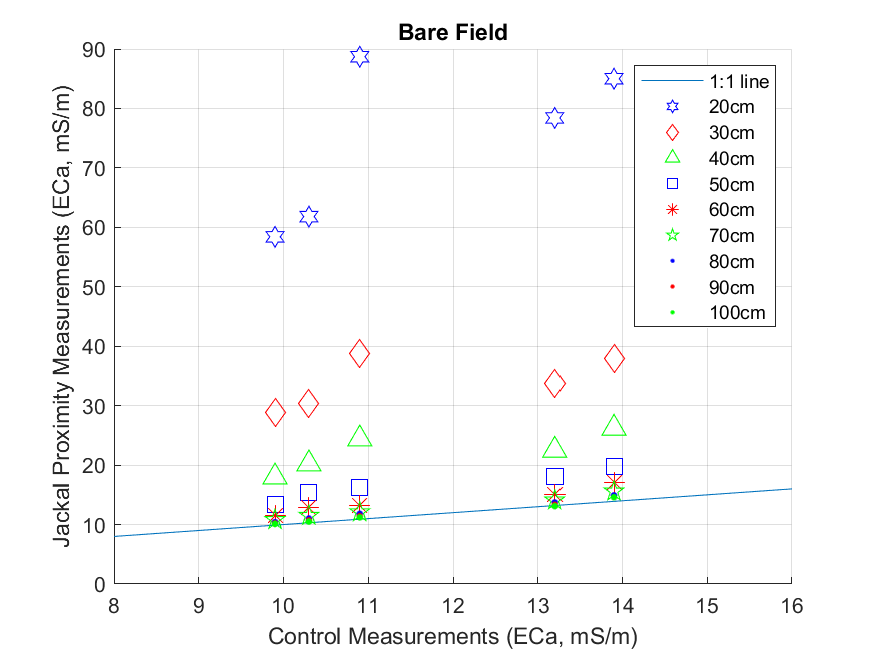}}
    \hfill
  \subfloat[\label{s3-b}]{%
        \includegraphics[width=0.33\linewidth]{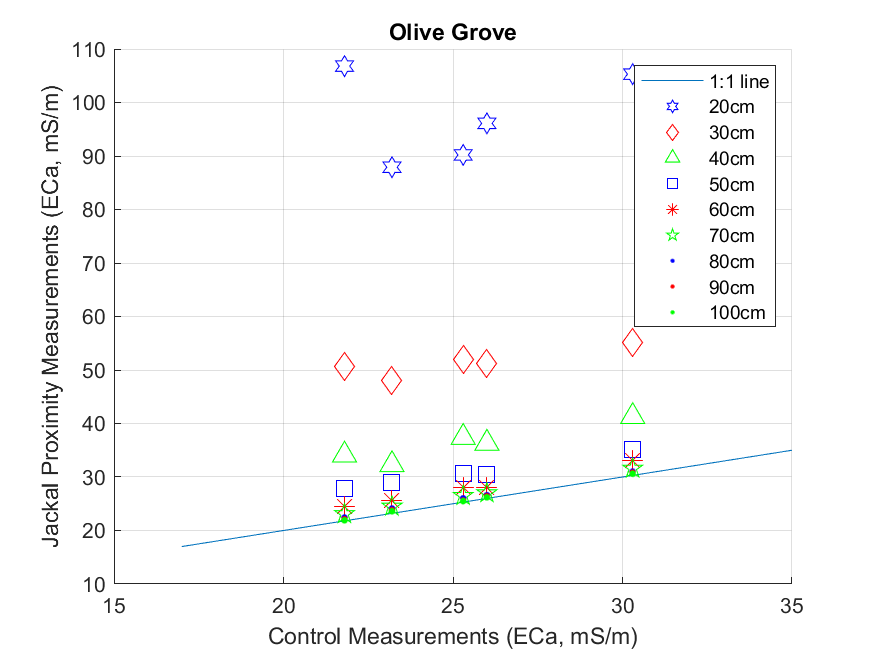}}
            \hfill
  \subfloat[\label{s3-c}]{%
        \includegraphics[width=0.33\linewidth]{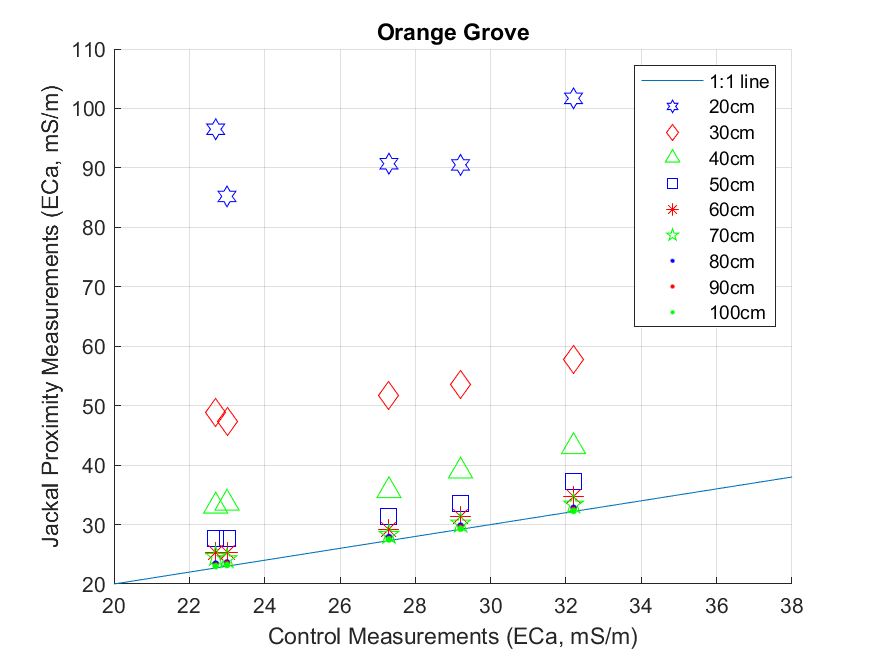}}
  \caption{1:1 control lines of soil conductivity measurements for the different robot-sensor distances at the three fields (best viewed in color).}
  \label{fig:controllines} 
  \vspace{5pt}
\end{figure*}

\begin{figure*}[!t]
    \centering
  \subfloat[\label{s4-a}]{%
       \includegraphics[width=0.25\linewidth]{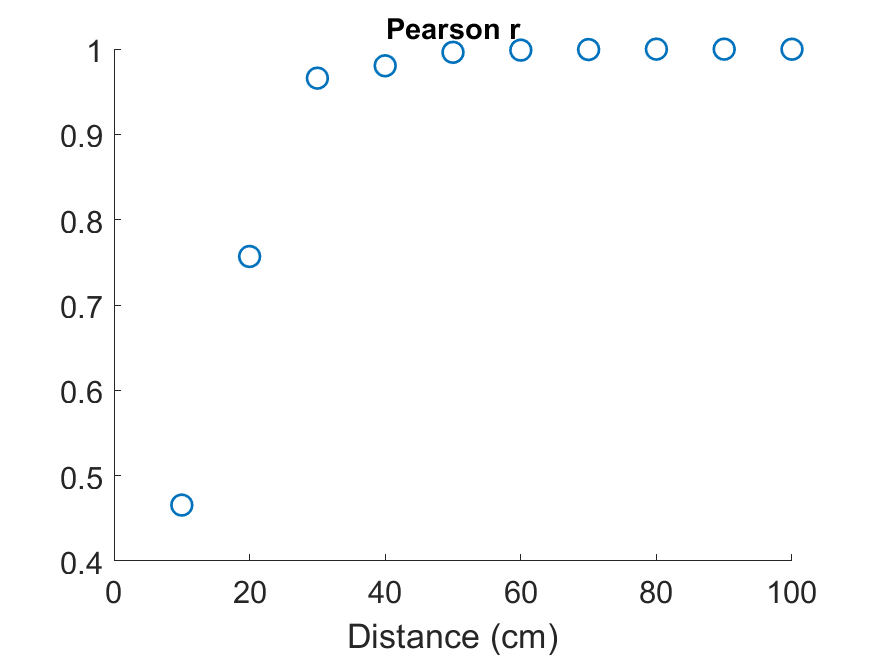}}
    \hfill
  \subfloat[\label{s4-b}]{%
        \includegraphics[width=0.25\linewidth]{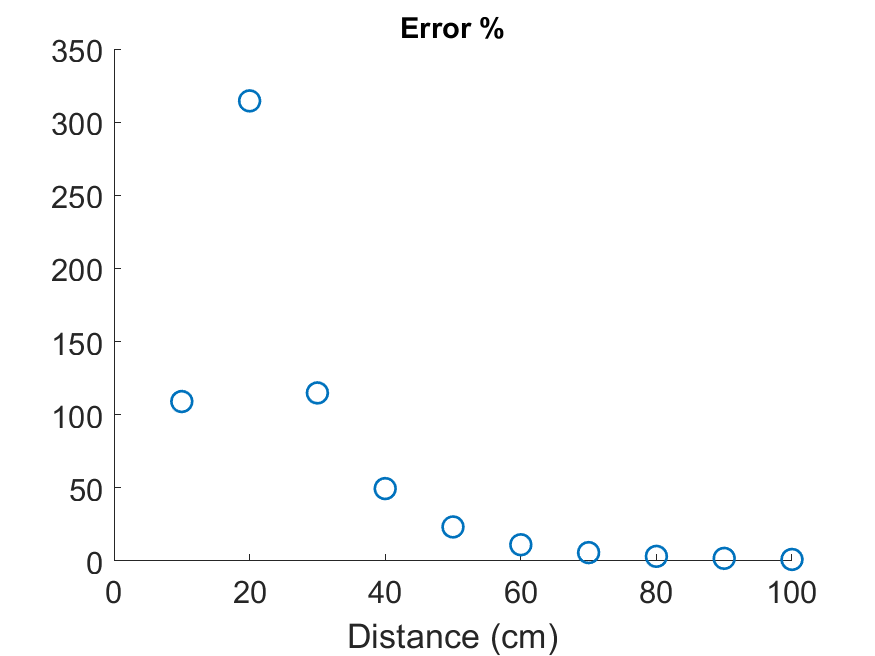}}
             \hfill
     \subfloat[\label{s4-c}]{%
        \includegraphics[width=0.25\linewidth]{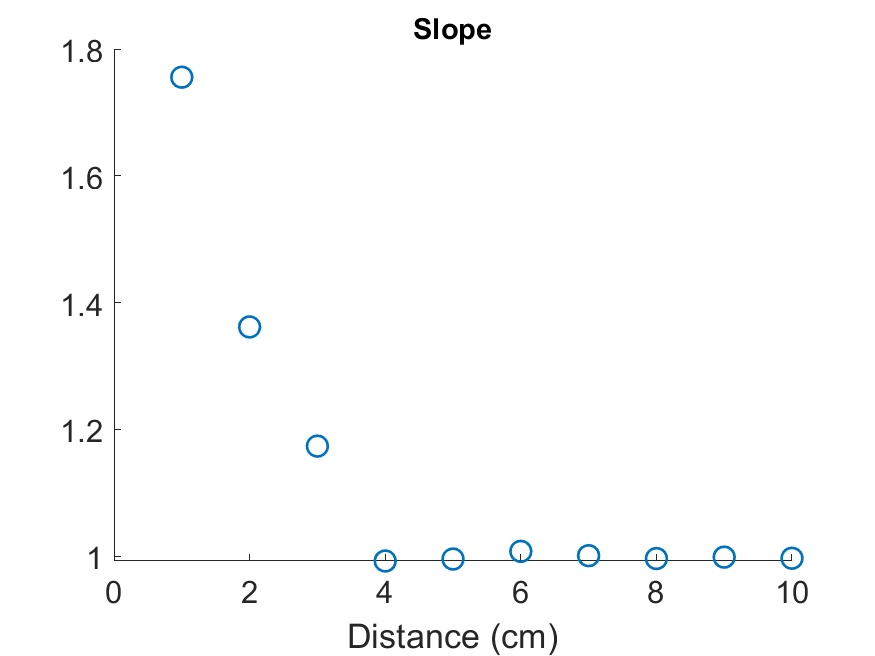}}
        \hfill
  \subfloat[\label{s4-d}]{%
        \includegraphics[width=0.25\linewidth]{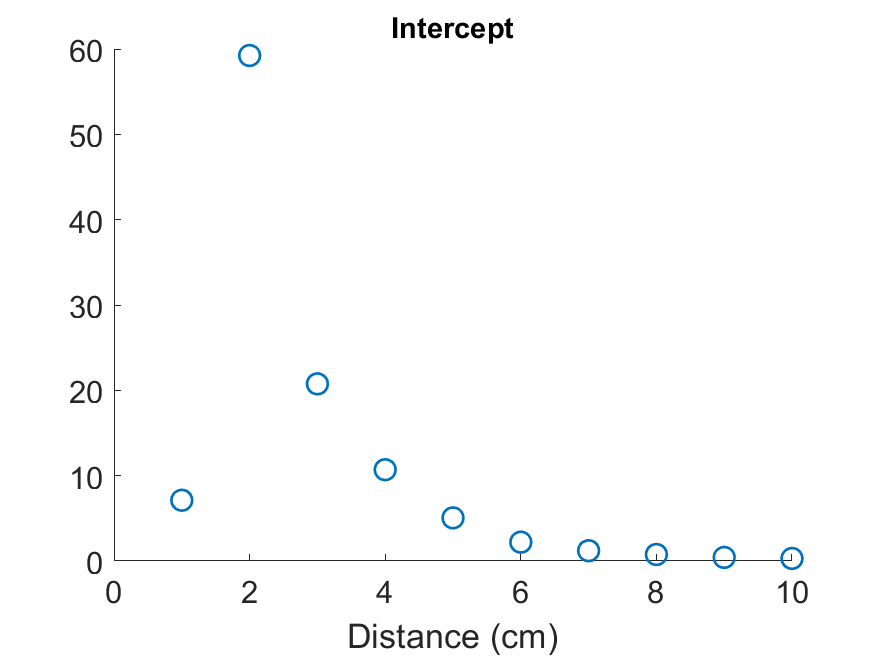}}
  \caption{Data analysis of the different robot-sensor distances with respect to the handheld-collected (baseline) data. (a) Pearson correlation test; (b) error percentage in measurements; (c) the slope value; and (d) the intercept value are depicted.}
  \label{fig:pearsonslopeinterect} 
  \vspace{5pt}
\end{figure*}

\begin{figure*}[!t]
    \centering
  \subfloat[\label{s5-a}]{%
       \includegraphics[width=0.20\linewidth]{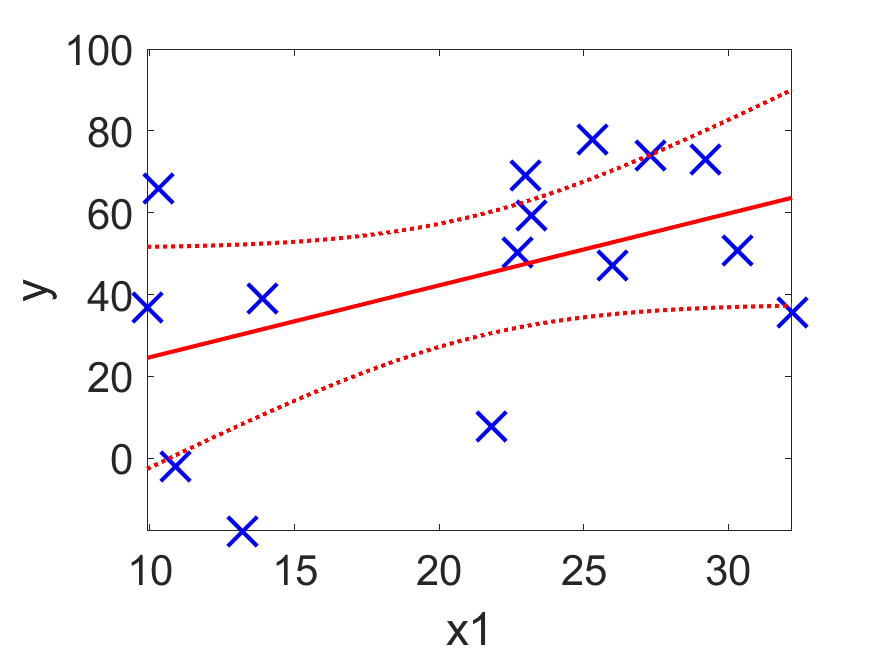}}
    \hfill
  \subfloat[\label{s5-b}]{%
        \includegraphics[width=0.20\linewidth]{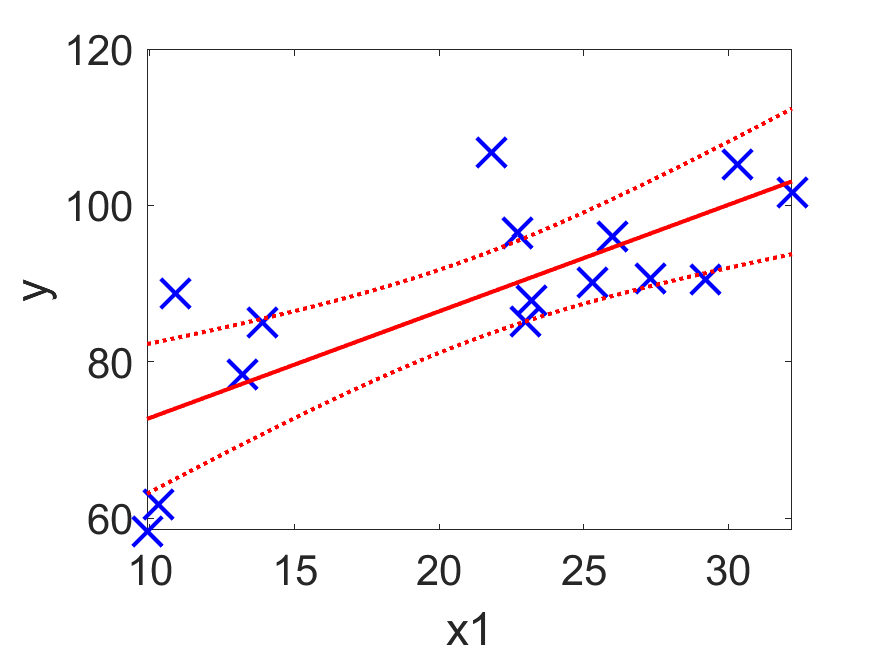}}
            \hfill
  \subfloat[\label{s5-c}]{%
        \includegraphics[width=0.20\linewidth]{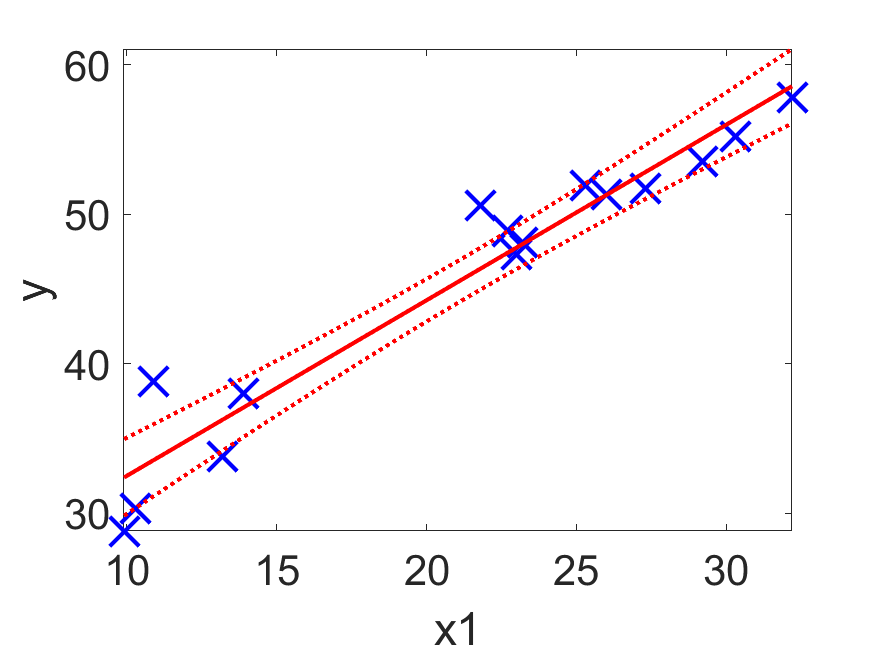}}
          \hfill
  \subfloat[\label{s5-d}]{%
        \includegraphics[width=0.20\linewidth]{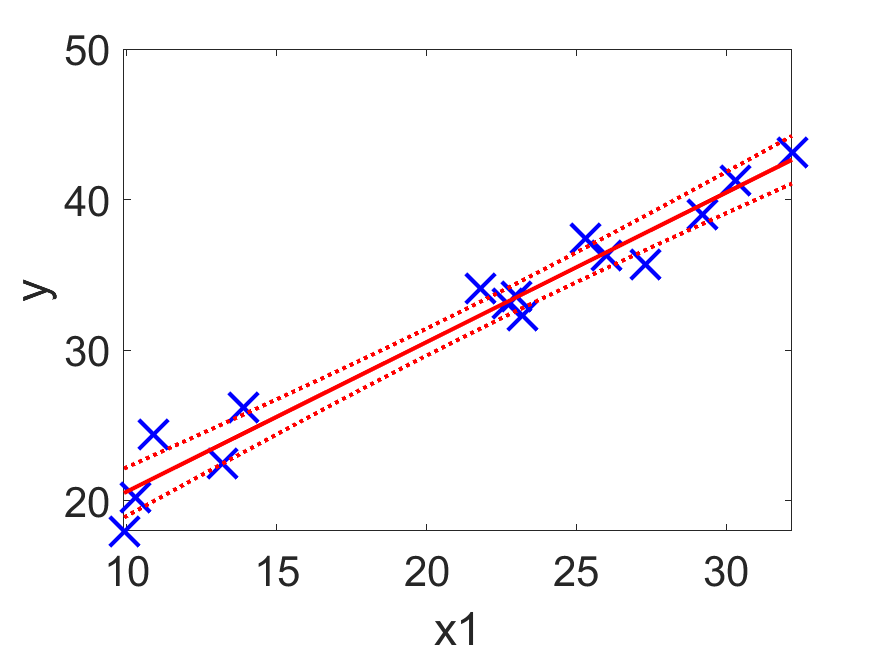}}
          \hfill
      \subfloat[\label{s5-e}]{%
    \includegraphics[width=0.20\linewidth]{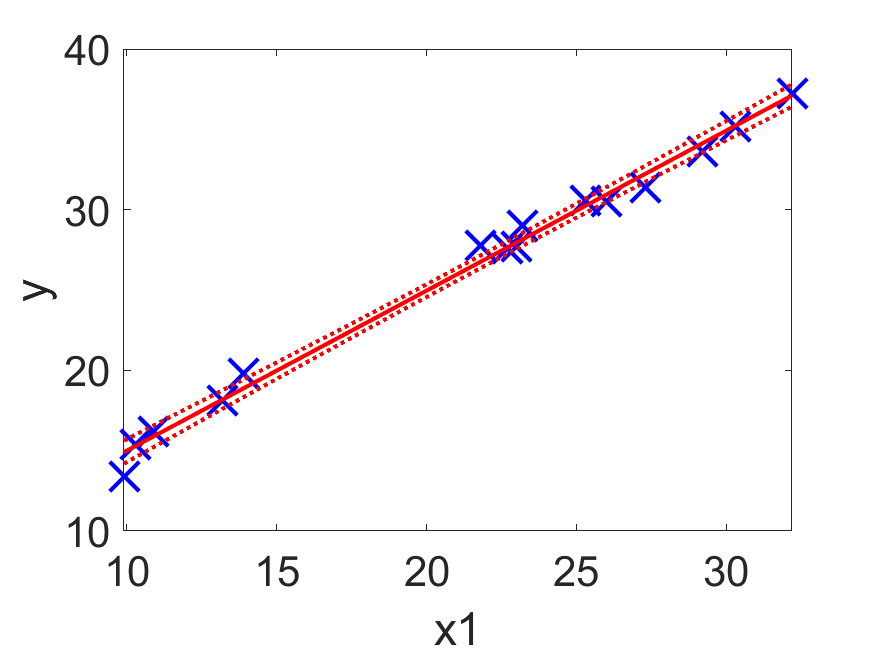}}
\\
    \subfloat[\label{s5-f}]{%
       \includegraphics[width=0.20\linewidth]{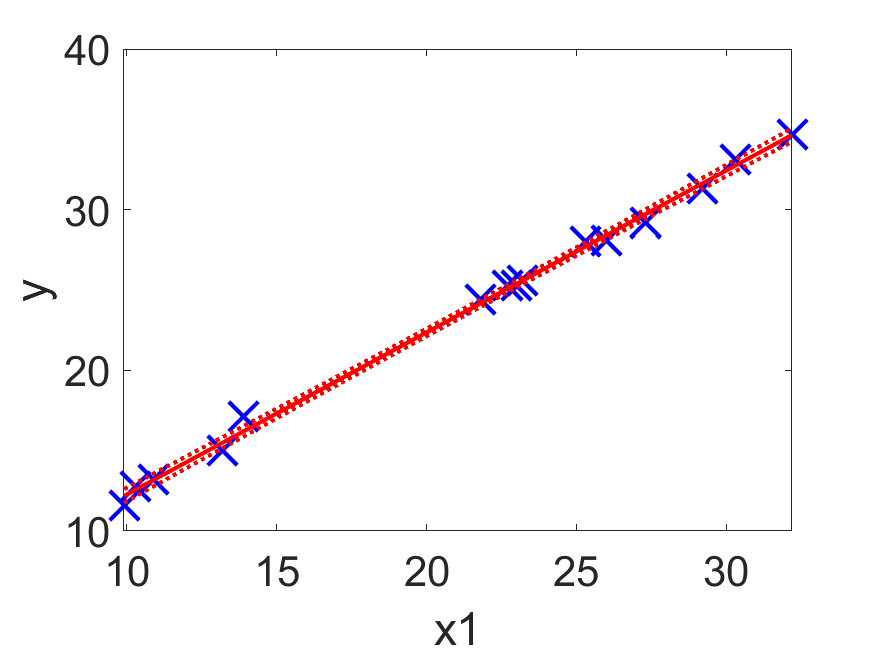}}
    \hfill
  \subfloat[\label{s5-g}]{%
        \includegraphics[width=0.20\linewidth]{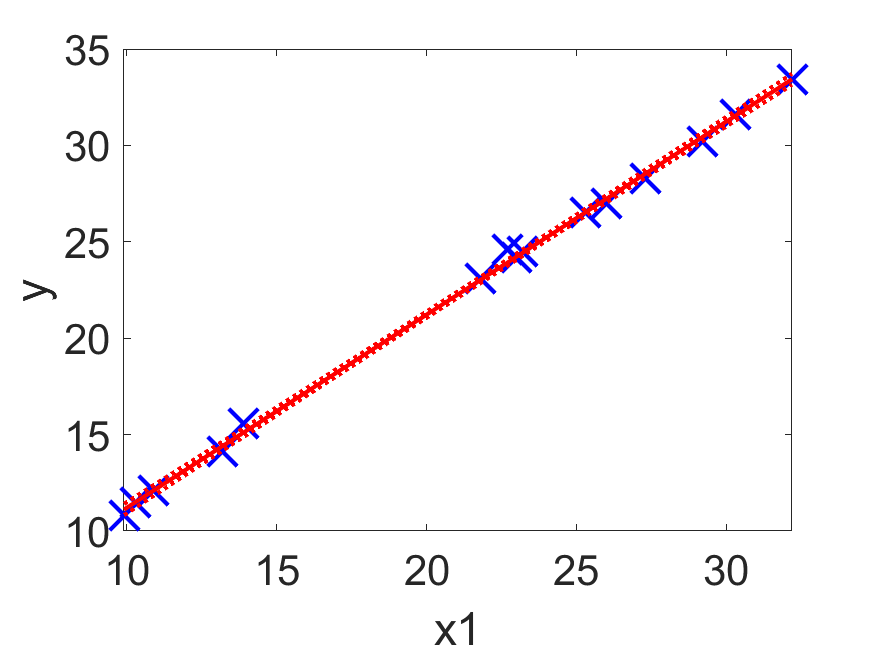}}
            \hfill
  \subfloat[\label{s5-h}]{%
        \includegraphics[width=0.20\linewidth]{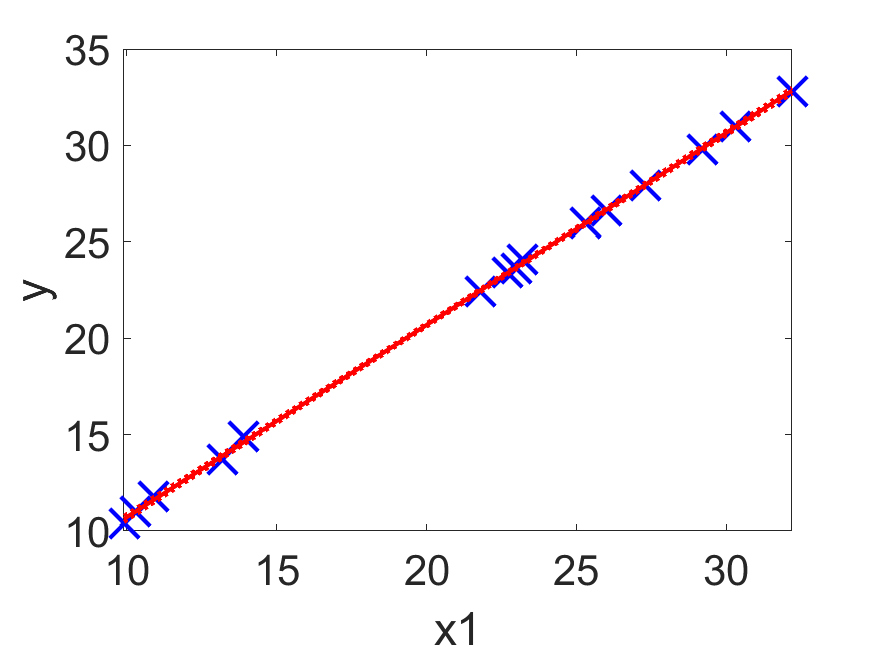}}
          \hfill
  \subfloat[\label{s5-i}]{%
        \includegraphics[width=0.20\linewidth]{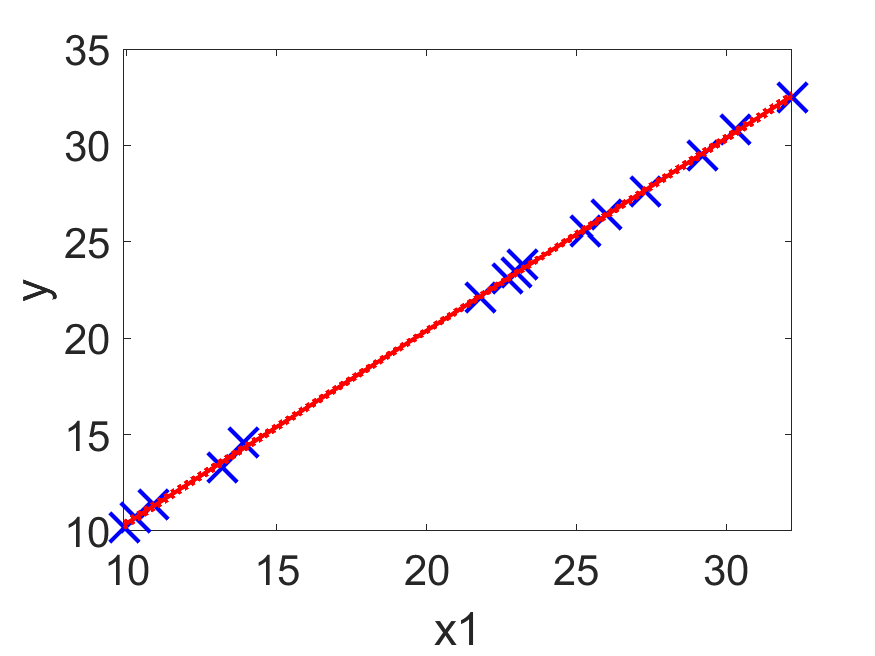}}
          \hfill
      \subfloat[\label{s5-j}]{%
    \includegraphics[width=0.20\linewidth]{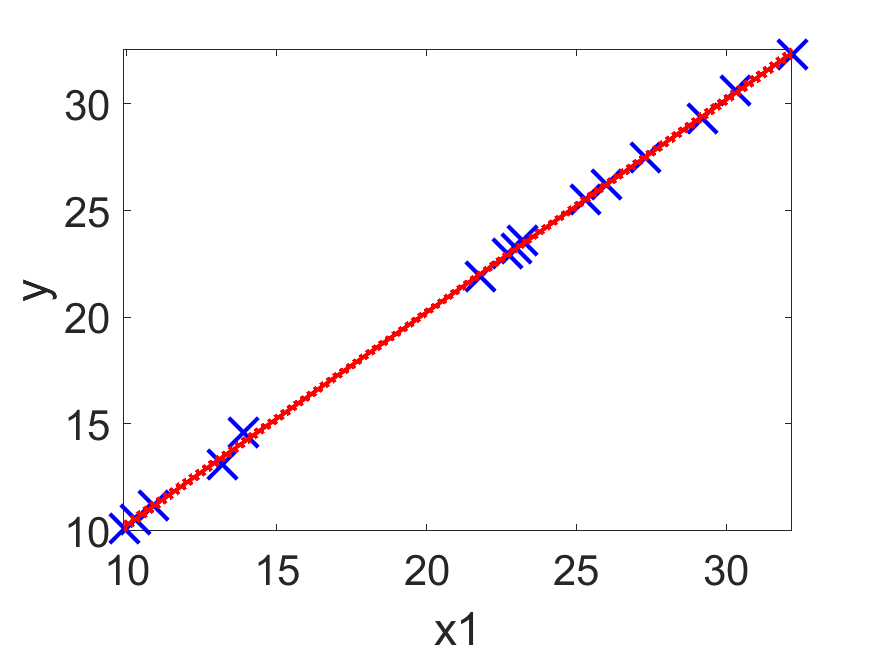}}
  \caption{Linear regression graphs of the captured conductivity data with respect to the baseline values, for evaluating the robot interference in the EMI measurements. Panels (a)--(j) correspond to each of the cases in the range $[10,100]\;cm$ of robot-sensor distance, respectively. Upper and lower bounds of the fit are depicted with continuous curves (in red).}
  \label{fig:linearregression} 
\end{figure*}

\subsubsection*{\textbf{Key Findings}}
Results of this first experimental benchmark reveal that the optimal placement of the EMI sensor has a lower threshold of $d_b=40\;cm$ away from the UGV body chassis. Specifically, numerical values reported in Table~\ref{tab:prwthmera} show that the saturation is notable in the measurements when the sensor is less than $40\;cm$ away from the robot's body. Shorter distances ($\{10,20,30\}\;cm$) exhibit both significantly higher mean values but also more variability (i.e. higher one-standard deviation) across all three fields. Especially the $d_b=10\;cm$ case exhibits excessive interference and clear evidence of saturation (the bare field depicts this most clearly). In contrast, larger $d_b$ values exhibit smaller interference as it can be readily verified by the reported means and one-standard deviations that converge to the baseline values. In addition, it can be observed that after some threshold distance, means and one-standard deviations do not vary significantly. In this work, this upper threshold is selected at $d_b=70\;cm$. 

Another interesting observation concerns the cross-field variability. The olive and citrus orchards that are regularly irrigated yield similar values across all tests, whereas the bare field (not irrigated) has lower ECa values (as expected). 
However, the measured values appear to converge faster (i.e. in shorter distances) in the irrigated fields compared to the bare field. We can associate this finding with the fact that in irrigated fields the reported ECa values are a function of both actual soil electrical conductivity and electromagnetic interference from the robot, and the former dominates more rapidly as the probe distance increases. 
In contrast, in the bare field, the actual ECa level attains much lower values, hence readings are more susceptible to electromagnetic interference and larger distances appear to be required to dampen down the interference's effect. This finding can be a useful tool to correlate soil salinity over the same field before and after planting, and while growing.

We can further justify the selection of the lower and upper thresholds for $d_b$ by inspecting measurement linearity compared to the baseline (Fig.~\ref{fig:controllines}) and via a Pearson correlation test (Fig.~\ref{fig:pearsonslopeinterect}) and subsequent linear regression (Fig.~\ref{fig:linearregression}). 
According to the $1:1$ control lines in all three panels of Fig.~\ref{fig:controllines} that correspond to each tested field, the saturation is notable in the measurements when the sensor is less than $40\;cm$ away from the robot's body; obtained values for $30\;cm$ and below clearly deviate from the other cases, while obtained values of $70\;cm$ are getting closely clustered together and approaching the $1:1$ control line.\footnote{~Note that the case of $10\;cm$ has significant interference and widely varied values (at cases even negative as shown in Table~\ref{tab:prwthmera}) and is henceforth not shown in these graphs to improve visual clarity of the figure.} 

The case of $d_b=40\;cm$ appears to be the switching point. While it could be argued based on Table~\ref{tab:prwthmera} and Fig.~\ref{fig:controllines} that this case demonstrates high differentiation from the baseline (especially at the bare field), the Pearson correlation \modd{test} (Fig.~\ref{fig:pearsonslopeinterect}a) yields a value of $0.98$ for the distance of $40\;cm$, which approximates the value of 1 and thus can be considered as a constant linear offset. 
Slopes and linear regression results also provide additional supporting evidence for picking $d_b=40\;cm$ as the lower threshold for further analysis. It can also be readily verified that there are no significant differences from setting the probe further than $70\;cm$ away from the robot body, as obtained values appear to have stabilized very close to the respective baseline measurement, in all three fields. Considering that the further we place the probe the worse its terrain traversability capacity (see next section too), we deduce that $d_b=70\;cm$ is an appropriate upper threshold for further analysis. Finally, Fig.~\ref{fig:pearsonslopeinterect}d shows the intercept coefficient decrease as the probe distance increases which corroborates all previous observations. 

Given all these remarks, we validate that keeping the sensor at a certain distance and farther from the robot's body frame leads to a decrease of the included noise in the conductivity measurements because of electromagnetic interference and increases the linearity of the given readings with respect to the reference (handheld) measurements. As such, we select the fixed distances of $d_b\in\{40, 50, 60, 70\}\;cm$ to further evaluate the robot's traversability capacity and thus help select the optimal one. 

\subsection{Evaluation of Robot Platform Maneuverability though Gazebo Simulation}\label{seq:gazeboexperiments}

With the set of candidate sensor-to-body distance values been identified ($d_b\in\{40, 50, 60, 70\}\;cm$), the next step is to examine the robot's ability to move over various uneven terrain fields while carrying the probe without the probe colliding with the ground as the robot negotiates dips and bumps. 
In general, each platform configuration may have a distinct effect on the robot's traversability capacity, sensor stability\add{,} and thus sensor readings. In an effort to make this process scalable and generalizable, we develop a realistic simulation environment and test robot traversability and probe oscillations for different sensor placement configurations over varied sets of emulated terrains.

\subsubsection*{\textbf{Experimental Procedure}} 
\begin{figure}[!t]
    \centering
  \subfloat[\label{sfig:agisoft}]{%
       \includegraphics[height=3cm]{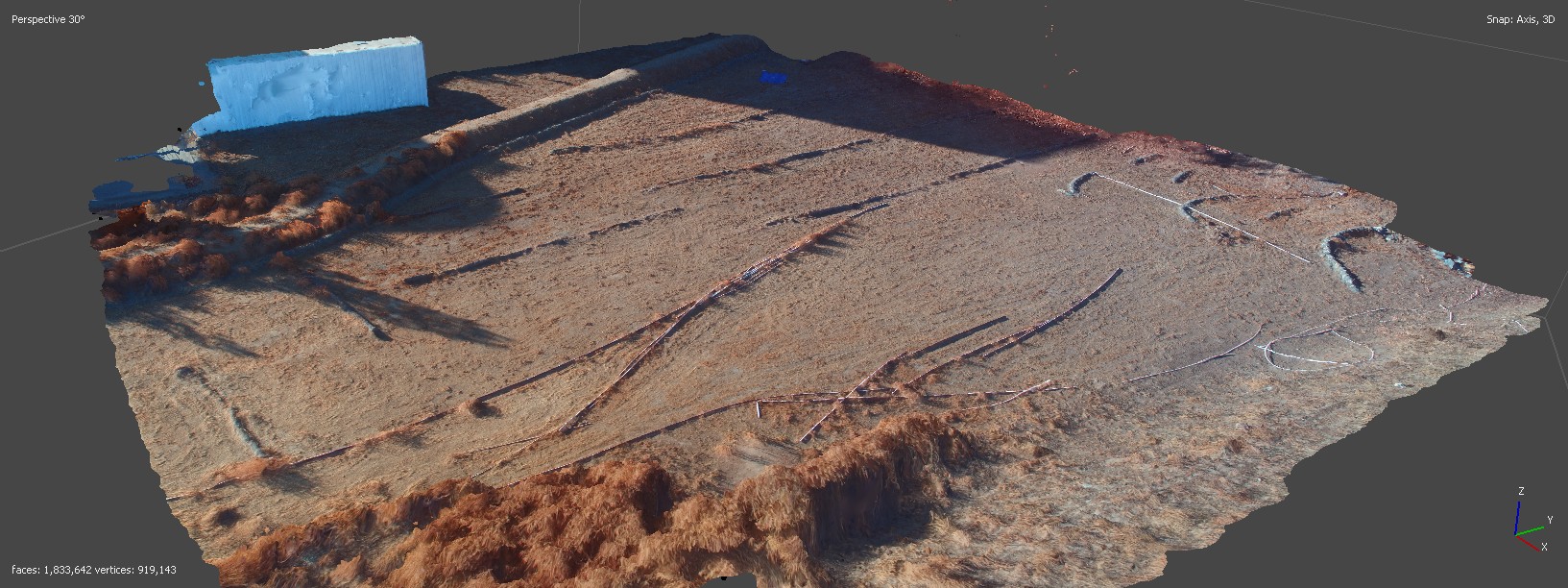}}
    \hfill
    \subfloat[\label{sfig:setups}]{%
        \includegraphics[height=3cm]{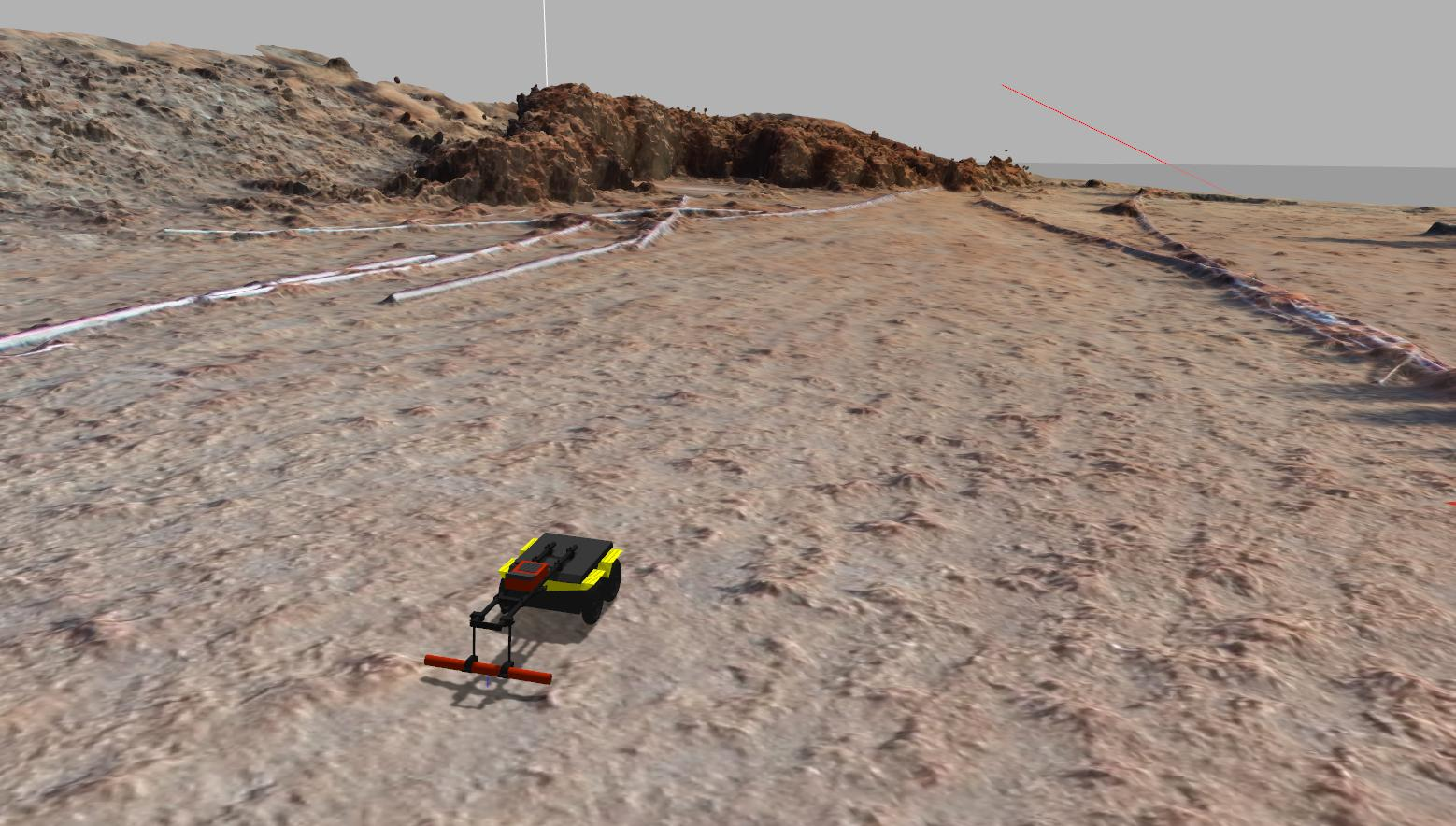}}
        \vspace{-3pt}
  \caption{Instances of the simulated environment created in this work. (a) Final 3D mesh in the Agisoft Metashape software. (b) The simulated Jackal robot equipped with the designed sensor platform spawned in the simulated environment.}
  \label{fig:agisoftresults} 
\end{figure}

We employed the Gazebo Robotics simulator~\cite{koenig2004design} and considered three key steps: 1) generation of 3D model maps based on aerial imagery, 
2) generation of the robot and sensor models, and 
3) set up of the simulated experiments and measured variables. The first step allows for the generation of a realistic emulated environment based on real imagery data. We used the photogrammetric software Agisoft Metashape to create a 3D world model based on aerial imagery data that were captured just south of the Center for Environmental Research \& Technology (CERT) at the University of California, Riverside (33$^\circ$59'59.563267'' N, -117$^\circ$20'5.769141'' E). By using 97 aerial photos obtained from different positions above the CERT field, we applied photo alignment and stitching through the Agisoft Metashape libraries and generated a colored pointcloud of 25904 points. In addition, by enhancing the sparse pointcloud with more correspondences from the stitching process, a dense pointcloud was generated containing $\simeq$ 9 million points; the latter was then transformed into a 3D mesh model by applying point interpolation. Figure~\ref{sfig:agisoft} shows the final form of the 3D mesh of the generated world, inside the Agisoft Metashape interface, and Fig.~\ref{sfig:setups} depicts the integration of the world into the Gazebo simulator along with the simulated real-scale robot.

\begin{figure}[!t]
\vspace{-3pt}
    \centering
  \subfloat[\label{sfig:systemsim}]{%
       \includegraphics[height=2.5cm]{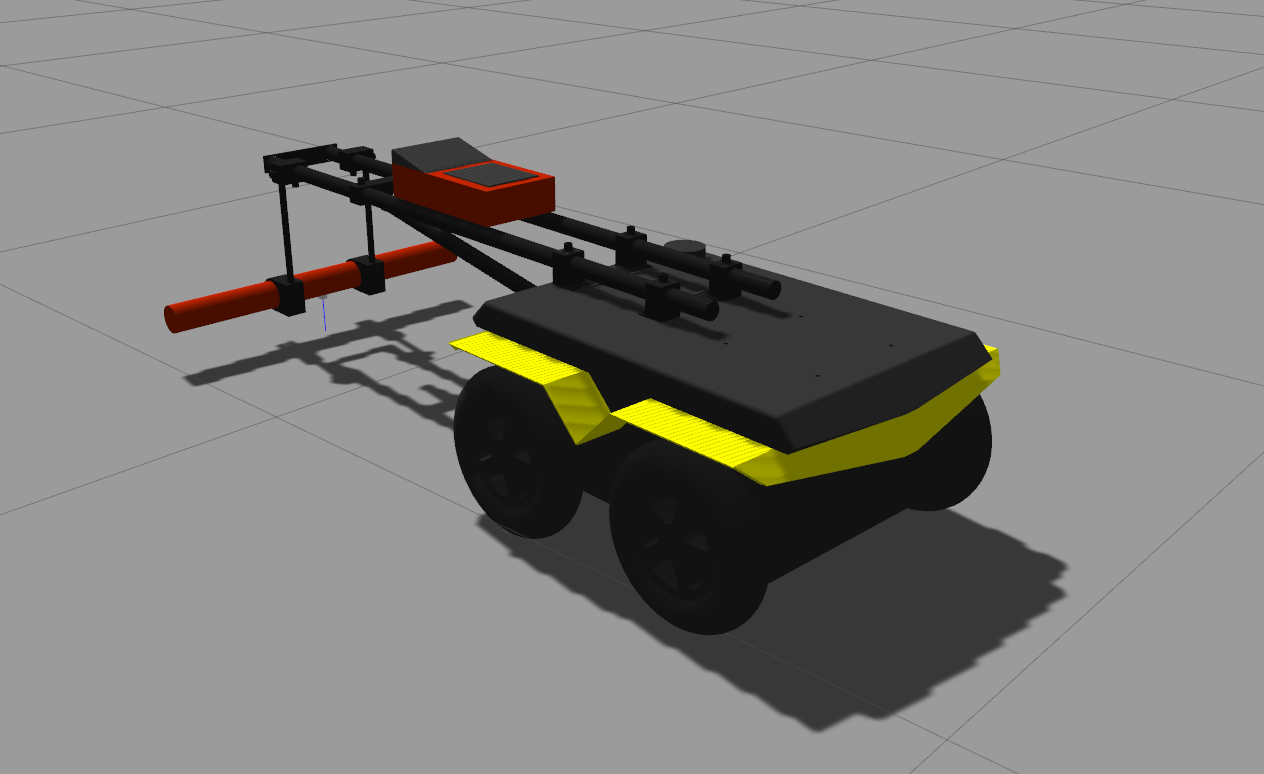}}
       \hfill
\subfloat[\label{sfig:blender}]{%
       \includegraphics[trim={3cm, 0cm, 3cm, 3cm},clip,height=2.5cm]{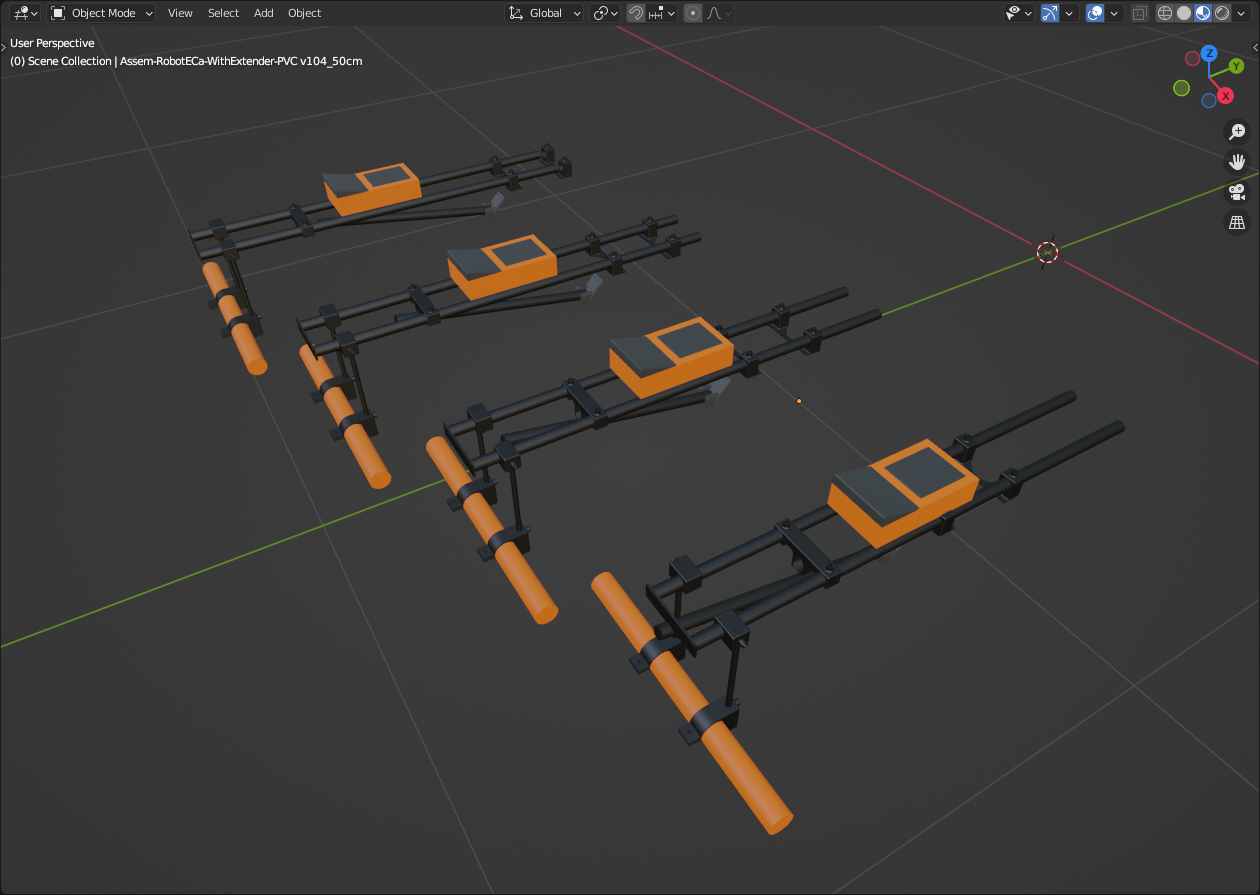}}
    \hfill
\subfloat[\label{sfig:oscilation}]{%
        \includegraphics[trim={1cm, 0cm, 3cm, 1cm},clip,height=3.7cm]{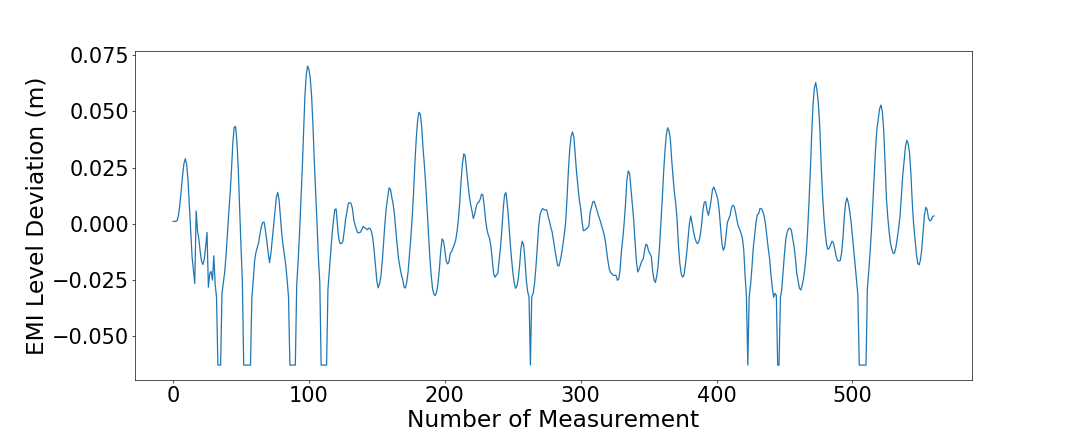}}

\vspace{-3pt}
\caption{(a) Detailed view on the simulated robot with the sensor mounted. (b) Generated 3D models of the EMI platform in the configuration. (c) Time series chart of level deviations of the sensor during its random roam in the simulated environment. \label{fig:gazebosnapshots}}
\end{figure}

\begin{figure}[!t]
\vspace{-3pt}
    \centering
  \subfloat{%
       \includegraphics[trim={0.5cm, 4cm, 0cm, 2cm},clip,height=4.2cm]{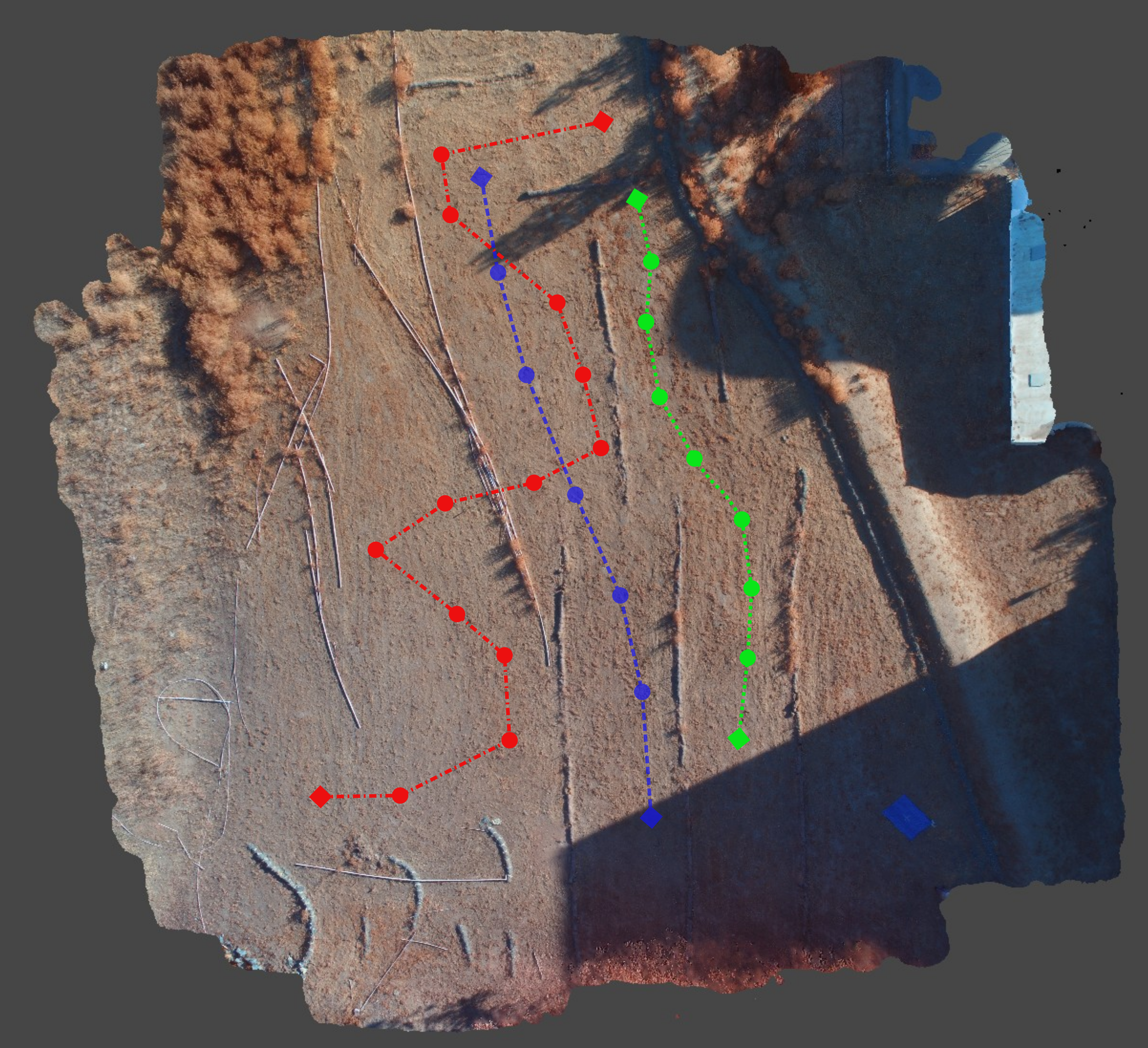}}
\vspace{-3pt}
  \caption{Planned trajectories for testing robot traversability in the simulated Gazebo world. The blue (6-node) trajectory lies on a smooth region of the map, the green (8-node) is on an uneven/rocky land area of the map, and lastly the red (13-node) contains a longer trajectory over mixed type terrain (best viewed in color).}
  \label{fig:gazebotrajectories} 
  \vspace{-3pt}
\end{figure}

A Gazebo model for the Clearpath Robotics Jackal robot employed herein is already publically available.\footnote{~\url{http://wiki.ros.org/Robots/Jackal}} We developed a custom Gazebo model for the GF CMD-Tiny sensor and integrated into the main robot model. Figure~\ref{fig:gazebosnapshots} depicts these models. 
Four separate Gazebo models of the sensor were generated (Fig.~\ref{sfig:blender}), each corresponding to one of the considered probe distances $d_b\in\{40, 50, 60, 70\}\;cm$. The sensor's height off the ground was adjusted to be $d_h\in\{6, 11\}\;cm$, which approximates the average height of sensor placement during handheld continuous measurements. The simulated Jackal model weighs $17\;kg$, similar to the real one, and the sensor platform with the onboard CMD-Tiny sensor is at $4\;kg$. 
To evaluate the GF CMD-Tiny sensor oscillations in the $z$-axis during the Jackal's movement, a ranging plugin was also developed in Gazebo, to provide continuous information about the distance of the EMI sensor from the ground. The Gazebo plugin publishes continuously the ranges of the sensor body to the ground through a $sensor\_msgs/LaserScan$ ROS topic, thus allowing us to capture the deviation data in real-time (one such example is shown in Fig.~\ref{sfig:oscilation}). 

Note that the simulated setup features the same ROS libraries as the real robot does, and it can be teleoperated or commanded to move to a goal pose (i.e. position and orientation) in the local frame, as it has an integrated RTK-GNSS antenna as well.

To evaluate the rigidity and robustness of the designed platform we conducted three independent simulated experiments inside the emulated field in the Gazebo world for the selected platform configurations (parameters $d_b$ and $d_h$). In these experiments, the Jackal is commanded to follow specific trajectories of different difficulty terrain types, namely a straight path in a slightly planar area, a straight path in a more rocky area, and a complex path with mixed type terrain (Fig.~\ref{fig:gazebotrajectories}). During \add{the} trajectory following, the robot measures the vertical oscillations of the EMI sensor with respect to the ground. Each individual case is repeated for five times, thus giving rise to a total of 120 (simulated) experimental trials in this specific benchmark.

\subsubsection*{\textbf{Key Findings}}
Results of this second experimental benchmark are contained in Table \ref{tab:gazebotab}. First, we confirm (as anticipated) that the increase of the sensor distance relative to the robot body makes the robot less stable and in turn leads to increased probe oscillations when traversing uneven terrain. However, it turns out that the cases of $d_h=\{50, 60\}\;cm$ lead to very similar platform oscillations in terms of reported variance, which in fact are close to the shortest case of $40\;cm$ and more steady compared to the longest $70\;cm$ case. In more detail, and with reference to Table~\ref{tab:gazebotab}, the increase of the distance in the probe placement results in increased mean and variance of position deviations while moving, for both tested sensor height levels off the ground.

Mounting the sensor closer to the robot's body and keeping a shorter robot-sensor footprint, the robot is more stable in either smoother or more rough terrain types. For the $d_h=6\;cm$ case, placing the sensor at $d_b=\{50, 60\}\;cm$ achieves similar results by keeping $\simeq1\;cm$ of difference in standard deviation of the $70\;cm$ case. Also, even though $40\;cm$ is the less shaky solution overall, the cases of $d_b=\{50, 60\}\;cm$ have less than $\simeq0.5\;cm$ difference in standard deviation. Additionally, by comparing across experiments, the $50\;cm$ case appears to score less than $10^{-4}\;m^2$ of variance. These observations are consistent also in the case of $d_h=11\;cm$, by having the smallest sensor displacement when $d_b=40\;cm$ and with the cases of $d_b=\{50, 60\}\;cm$ performing equivalently. Based on these observations, we deduce that the cases of setting $d_b=\{50, 60\}\;cm$ may offer the best trade-off in terms of noise and steadiness compared with the shorter $40\;cm$ case, in which the probe is placed close to the robot's chassis and is affected more by the electromagnetic interference as shown in Section~\ref{sec:development}.

A second observation from these results concerns the sensor's height off the ground, $d_h$. Inspection of the results for different sensor height values $d_h$ in Table~\ref{tab:gazebotab}, we notice a similar oscillating behavior, whereby a shorter height (i.e. $d_h=6\;cm$) might be preferred to reduce variations from the ground. This becomes even more critical considering that, during real surveys, it is hard for a farm worker that uses the handheld sensor to keep the sensor consistently at a stable height off the ground while walking. For these reasons we chose to use $d_h=6\;cm$.

\begin{table*}[!t]
\caption{Measured sensor oscillations during simulated surveys in the Gazebo environment. \label{tab:gazebotab}}
\centering
\begin{tabular}{cccccccccc}
\toprule
                                 & \multicolumn{3}{c}{Smooth Land}                                                   & \multicolumn{3}{c}{Rocky Land}                                                    & \multicolumn{3}{c}{Mixed-terrain Land}                         
                    \\ \cline{2-10}
                              $d_b$ - $d_h$   & mean (cm)             & $\sigma$ (cm)           & \multicolumn{1}{c|}{variance ($cm^2$)} & mean (m)             & $\sigma$ (cm)           & \multicolumn{1}{c|}{variance ($cm^2$)} & mean (cm)             & $\sigma$ (cm)           & variance ($cm^2$)             
                    \\ 
                    \cline{2-10}
                                 &                      &                      &                                     &                      &                      &                                     &                      &                      &                        
                    \\
\multicolumn{1}{c|}{40cm - 6cm}  & -0.50              & 2.57               & \multicolumn{1}{c|}{6.61e-02}     & -0.51              & 3.15               & \multicolumn{1}{c|}{9.94e-02}     & -0.54              & 2.58               & \multicolumn{1}{c|}{6.65e-02} 
\\
\multicolumn{1}{c|}{50cm - 6cm}  & -0.60              & 3.13               & \multicolumn{1}{c|}{9.81e-02}     & -0.52              & 3.13               & \multicolumn{1}{c|}{9.77e-02}     & -0.68              & 2.80               & \multicolumn{1}{c|}{7.83e-02} 
\\
\multicolumn{1}{c|}{60cm - 6cm}  & -0.83              & 2.97               & \multicolumn{1}{c|}{8.81e-02}     & -0.49              & 3.24               & \multicolumn{1}{c|}{0.10}         & -0.79              & 3.02               & \multicolumn{1}{c|}{9.10e-02} 
\\
\multicolumn{1}{c|}{70cm - 6cm}  & 0.68               & 4.11               & \multicolumn{1}{c|}{0.17}         & -0.61              & 3.91               & \multicolumn{1}{c|}{0.15}         & -0.71              & 3.20               & \multicolumn{1}{c|}{0.10}     
\\
\midrule
                                 &                      &                      &                                     &                      &                      &                                     &                      &                      &                                 
\\
\multicolumn{1}{c|}{40cm - 11cm} & -0.84              & 2.97               & \multicolumn{1}{c|}{8.83e-02}     & -0.78              & 3.37               & \multicolumn{1}{c|}{0.11}         & -1.03              & 3.35               & \multicolumn{1}{c|}{0.11}     
\\
\multicolumn{1}{c|}{50cm - 11cm} & -1.12              & 3.39               & \multicolumn{1}{c|}{0.11}         & -1.03              & 4.07               & \multicolumn{1}{c|}{0.17}         & -1.25              & 3.66               & \multicolumn{1}{c|}{0.13}     
\\
\multicolumn{1}{c|}{60cm - 11cm} & -1.47              & 4.37               & \multicolumn{1}{c|}{0.19}         & -0.93              & 3.94               & \multicolumn{1}{c|}{0.16}         & -1.57              & 4.47               & \multicolumn{1}{c|}{0.20}     
\\
\multicolumn{1}{c|}{70cm - 11cm} & -1.23              & 3.69               & \multicolumn{1}{c|}{0.14}         & -1.62              & 4.76               & \multicolumn{1}{c|}{0.23}         & -1.50              & 4.04               & \multicolumn{1}{c|}{0.16}     
\\
\bottomrule
\end{tabular}
\end{table*}

%% file: fieldtests.tex
\subsection{Preliminary Feasibility Experimental Testing of Boundary Configurations}\label{sec:validation}

The last set of calibration tests for optimal sensor placement included validation of the preliminary feasibility of the robot-sensor setup to measure ECa continuously. Based on the aforementioned results, we elected to study in these validation experiments the two boundary cases of probe distance (i.e. $d_b=\{40, 70\}\;cm$), at a constant height of $d_h=6\;cm$.\footnote{~We wish to highlight at this point that the obtained results so far suggest that the configuration $\{d_b=60\;cm, d_h=6\;cm\}$ can serve as the optimal one in the context of this work. This is the configuration we test in field-level experiments in Section~\ref{sec:evaluation} that follows. However, for completeness purposes, and in an effort to better explain the effects of placing the sensor closer or further away from the robot on soil ECa measurements, we tested also the two boundary configurations (which are still viable in principle) but at a smaller-scale experimental setup compared to the field-level experimental setups in Section~\ref{sec:evaluation}.} By doing so, we can get a better understanding of the effects of placing the sensor closer or further away from the robot on \emph{continuous} ECa measurements.

\begin{figure}[!h] 
    \centering
  \subfloat[\label{jackalatcert}]{%
       \includegraphics[height=0.22\linewidth]{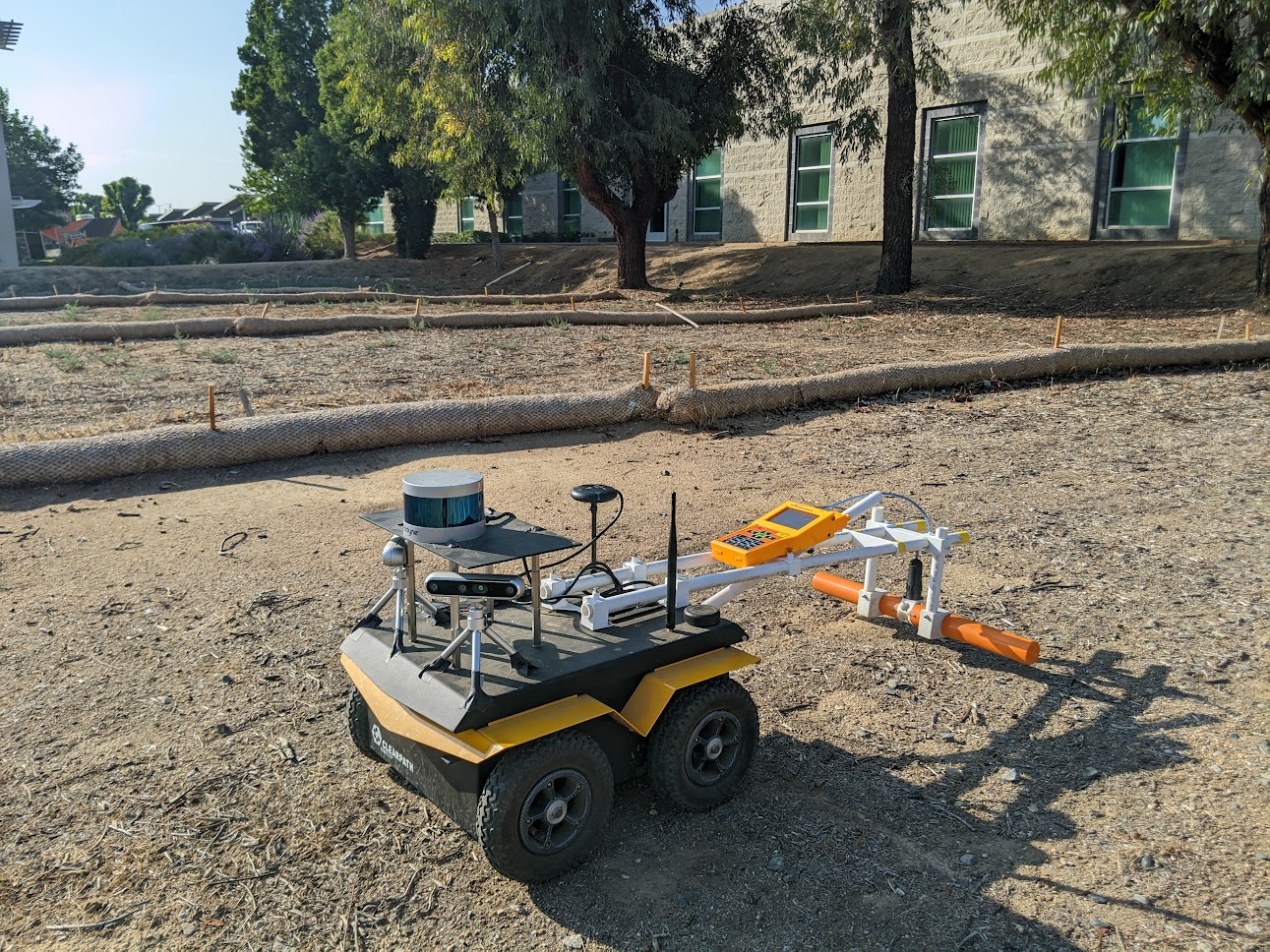}}
           \hfill
    \subfloat[\label{jackaldoingsurveycert}]{%
    \includegraphics[height=0.22\linewidth]{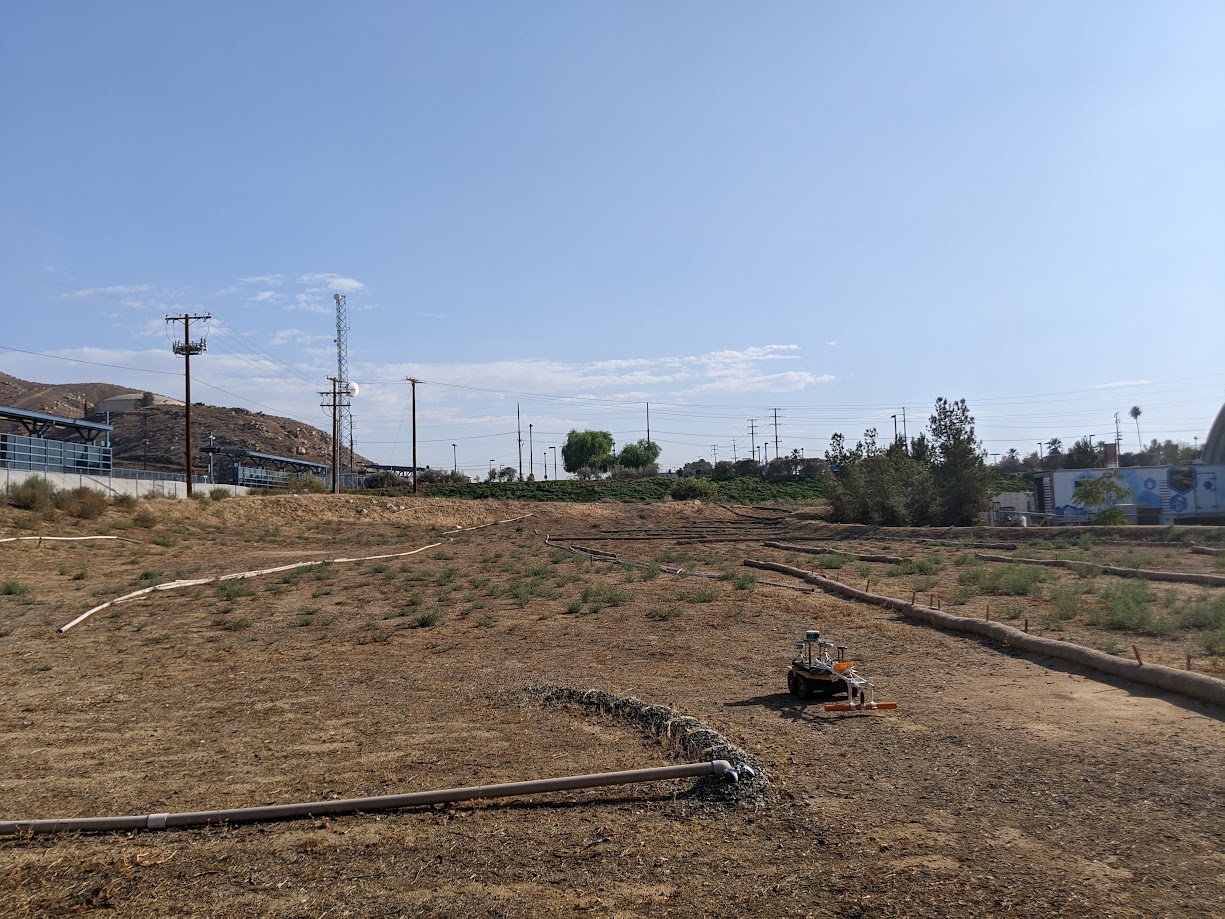}}
    \hfill
    \subfloat[\label{certtrajectories}]{%
    \includegraphics[trim={2cm, 0cm, 2cm, 2.5cm},clip,height=0.22\linewidth]{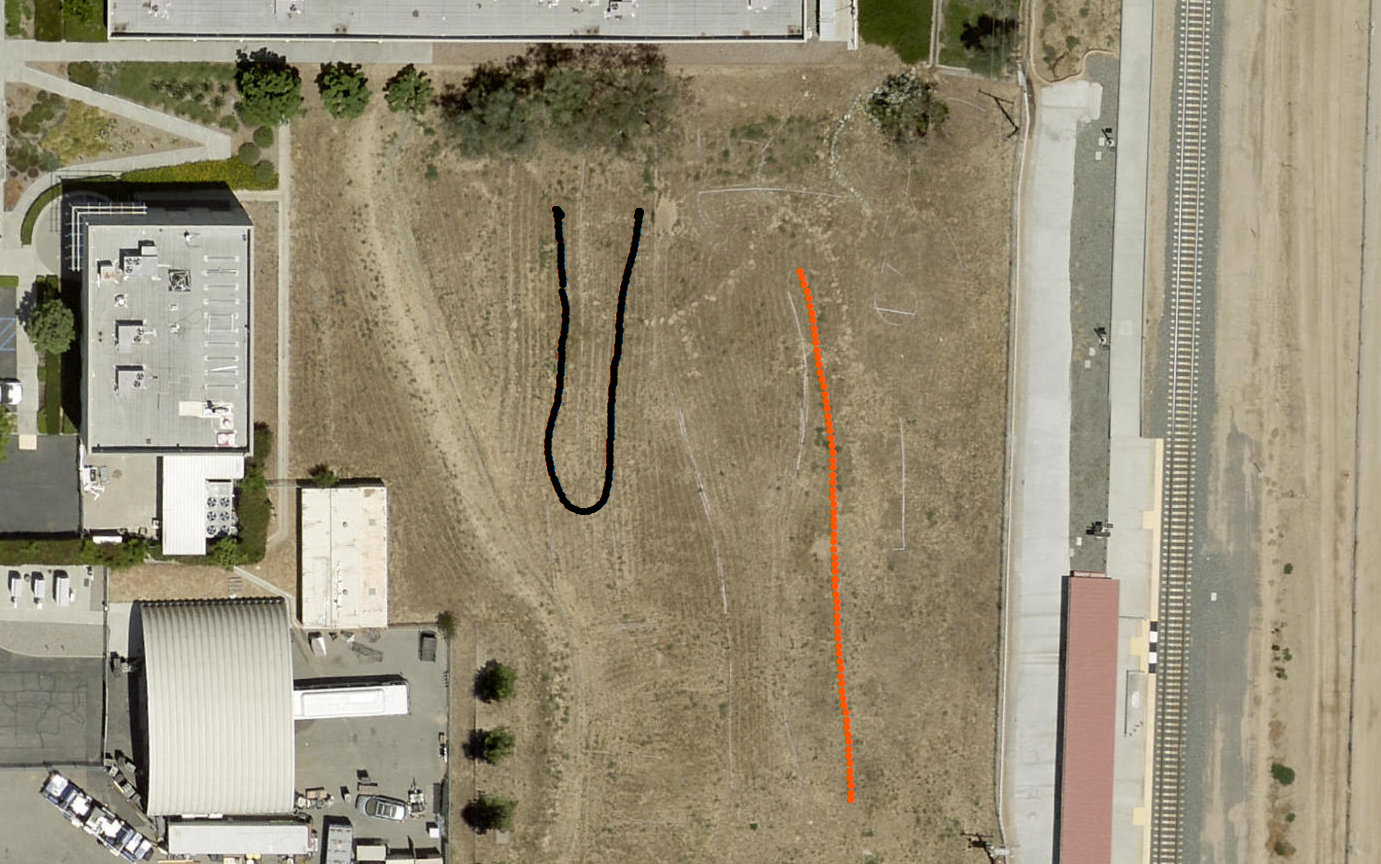}}
  \caption{(a) The Jackal robot equipped with the platform holding the CMD-Tiny instrument at the configuration of $d_b=50\;cm$ and $d_h=6\;cm$. (b) Instance of robot during the soil ECa data collection for validation. (c) Validation testing considered two distinctive trajectories, one following a straight line and another performing a U-shaped curve. \label{fig:validation}}
\end{figure}

\subsubsection*{\textbf{Experimental Procedure}} The validation experiments took place in the same field that was emulated for the aforementioned simulation testing. Without loss of generality, we considered two distinct cases: 1) a straight-line trajectory with $d_b=70\;cm$ and $d_h=6\;cm$, and 2) a U-shaped trajectory with $d_b=40\;cm$ and $d_h=6\;cm$. In each case we conducted three independent trials. The starting and end positions as well as intermediate waypoints were the same for each set of trials. The physical setup, experimental field, and the two types of trajectories are depicted in Fig.~\ref{fig:validation}. Manual data collections with the handheld sensor (three for each trajectory type) were also performed to serve as the baseline. In total, we conducted 12 experimental trials for validation in the bare field (six robotized and six manual). Obtained soil ECa measurements were georeferenced via RTK-GNSS in the robotized measurement and the sensor's embedded GPS in the manual measurements. Collected soil ECa data were used to generate a custom raster of the surveyed area; then we used kriging interpolation through exponential semivariogram to obtain the ECa map which was embedded onto satellite imagery via the ESRI ArcGIS 10.8.2 software. 

\subsubsection*{\textbf{Key Findings}}

\begin{figure}[!t]
    \centering
    \subfloat[\label{sf:plotcertstraight}]{%
       \includegraphics[trim={1cm, 0cm, 2.5cm, 1cm},clip,width=0.32\linewidth]{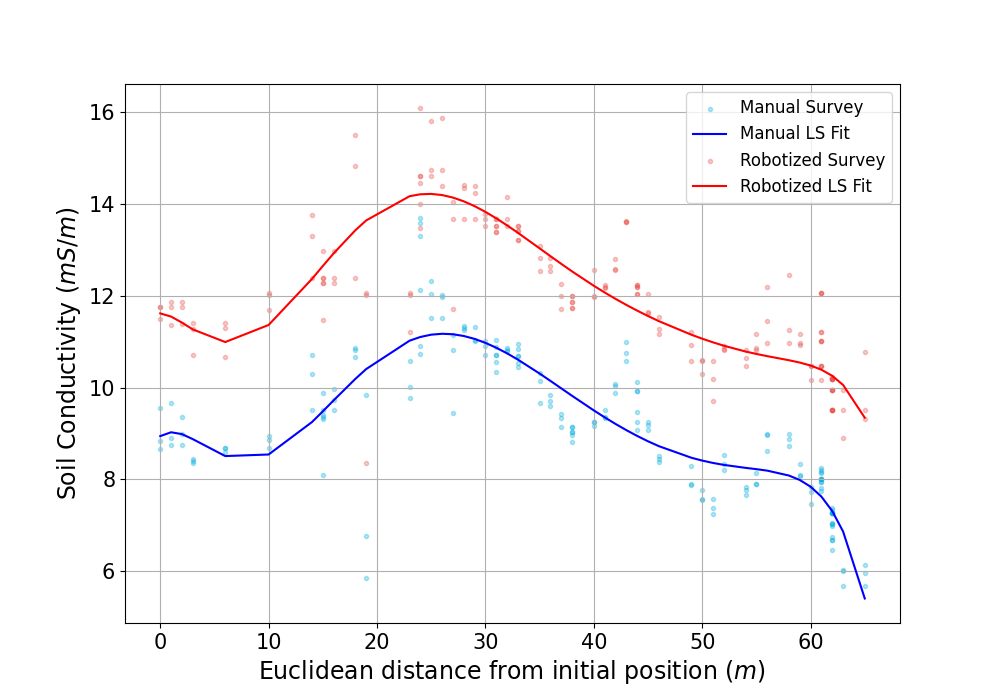}}
    \hfill
  \subfloat[\label{sf:handheld_arcgis_straight}]{%
       \includegraphics[trim={0cm, 0cm, 1.5cm, 0cm},clip,width=0.33\linewidth]{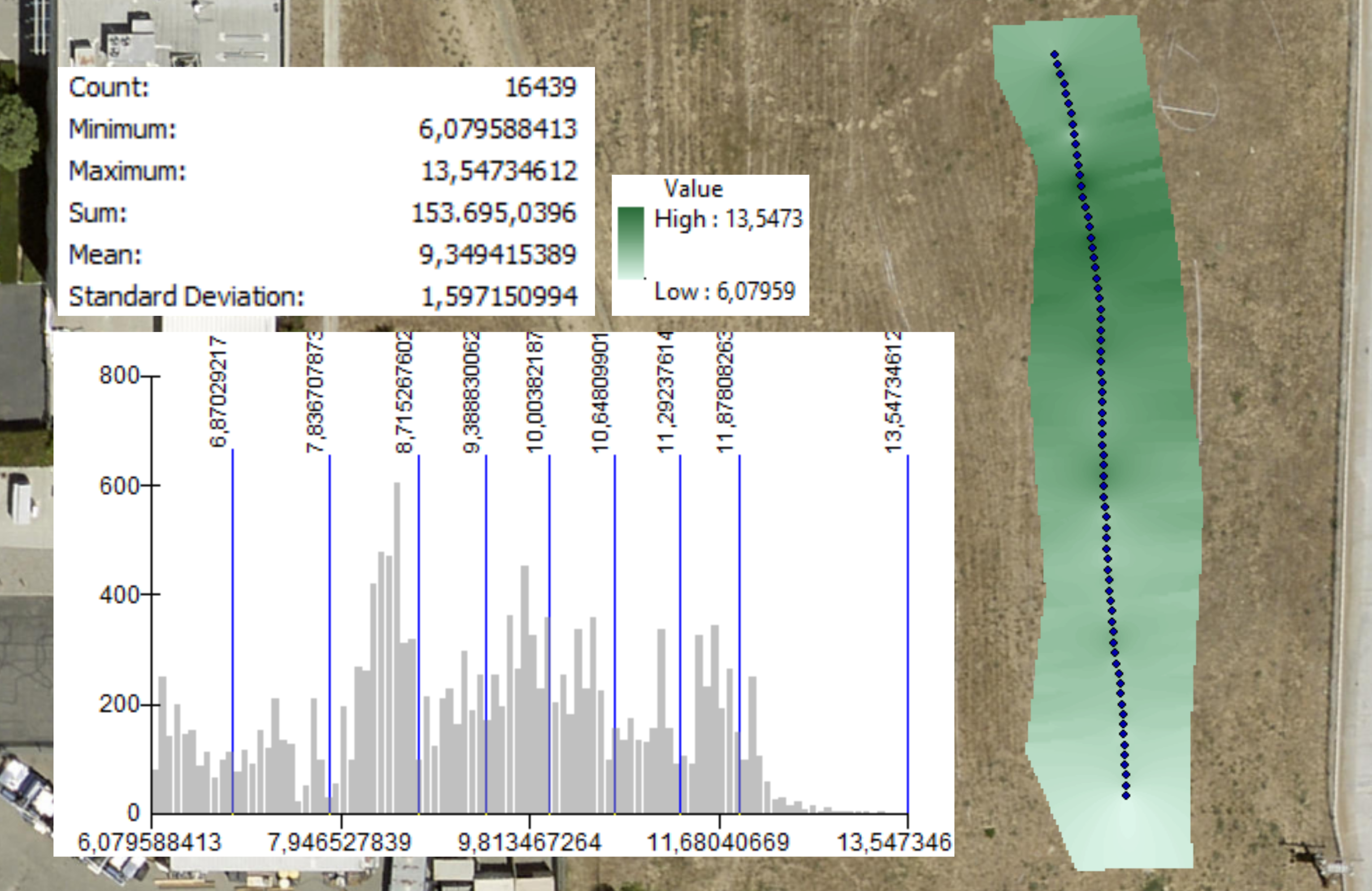}}
           \hfill
      \subfloat[\label{sf:jackal_arcgis_straight}]{%
    \includegraphics[trim={0cm, 0cm, 1.5cm, 0cm},clip,width=0.33\linewidth]{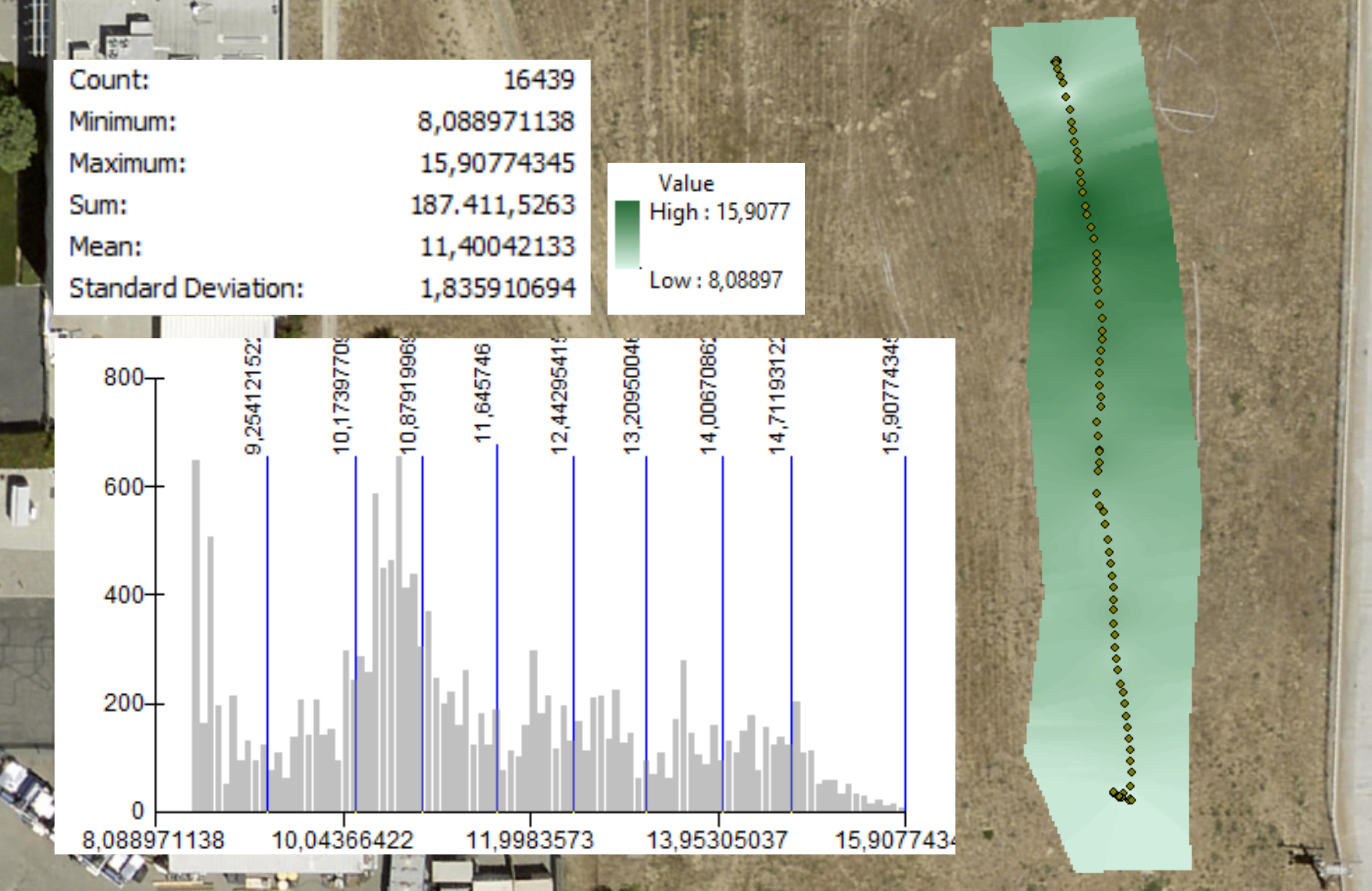}}
\caption{(a) The soil conductivity curves of the straight-line trajectory case in the bare field. Fitted graphs of both measurement curves correspond to an 8th grade least squares fit. (b)-(c) Soil ECa maps corresponding to manually-collected (handheld) and robotized-collected data, respectively, computed by applying kriging interpolation through exponential semivariogram. Each panel also contains the value-based color scale, the map statistics, and the histogram of the conductivity values.\label{fig:barevalidation70}}
\end{figure}

\begin{figure}[!t]
    \centering
    \subfloat[\label{certturn_mean}]{%
        \includegraphics[trim={1cm, 0cm, 2.5cm, 1cm},clip,width=0.32\linewidth]{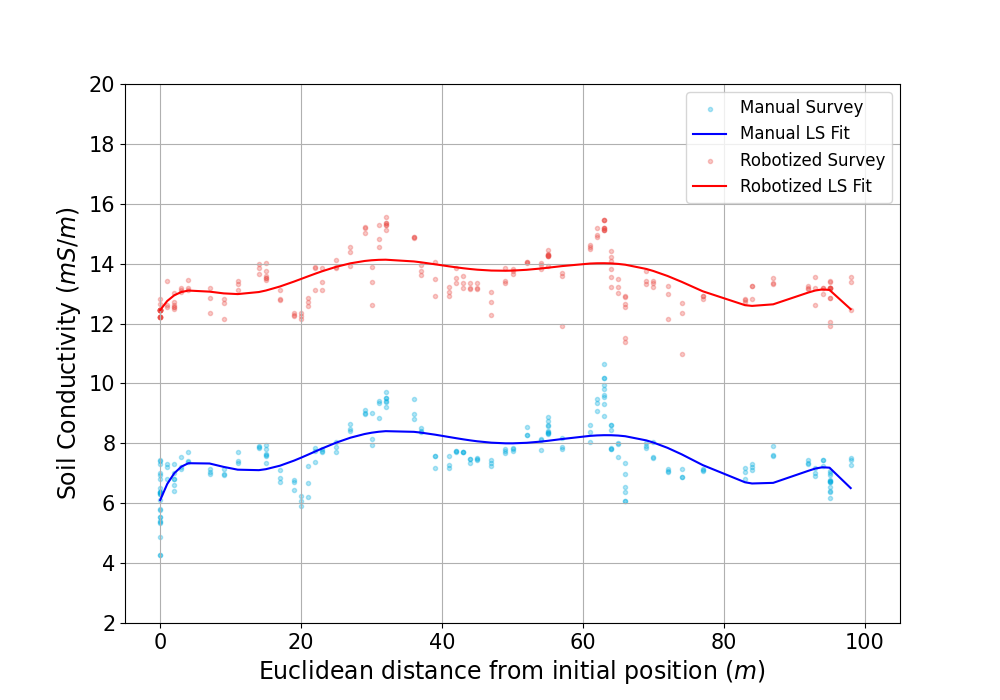}}
    \hfill
    \subfloat[\label{sf:handheld_arcgis_curved}]{%
       \includegraphics[trim={0cm, 0cm, 1cm, 0cm},clip,width=0.33\linewidth]{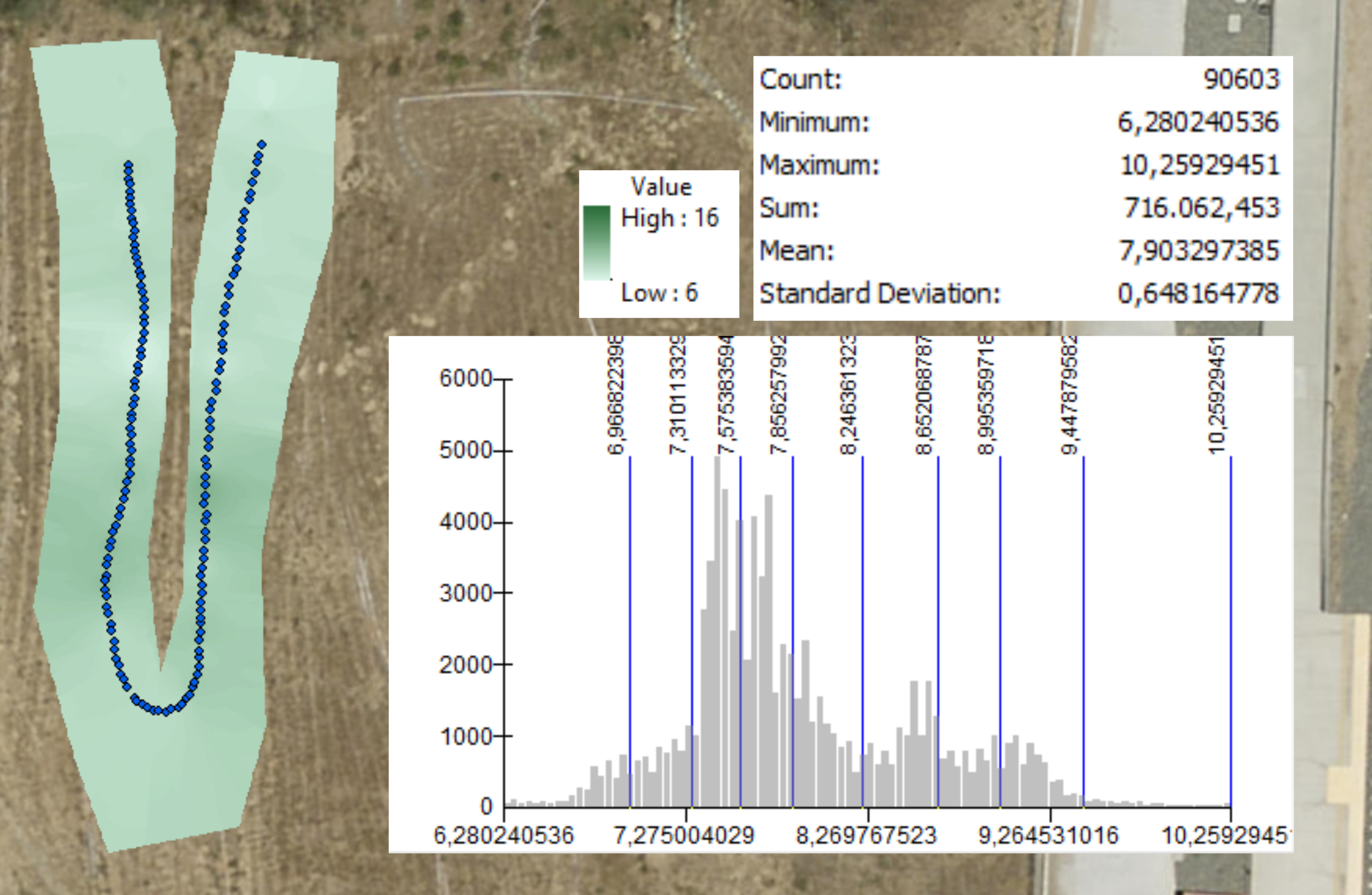}}
    \hfill
    \subfloat[\label{sf:jackal_arcgis_curved}]{%
        \includegraphics[trim={0cm, 0cm, 1cm, 0cm},clip,width=0.33\linewidth]{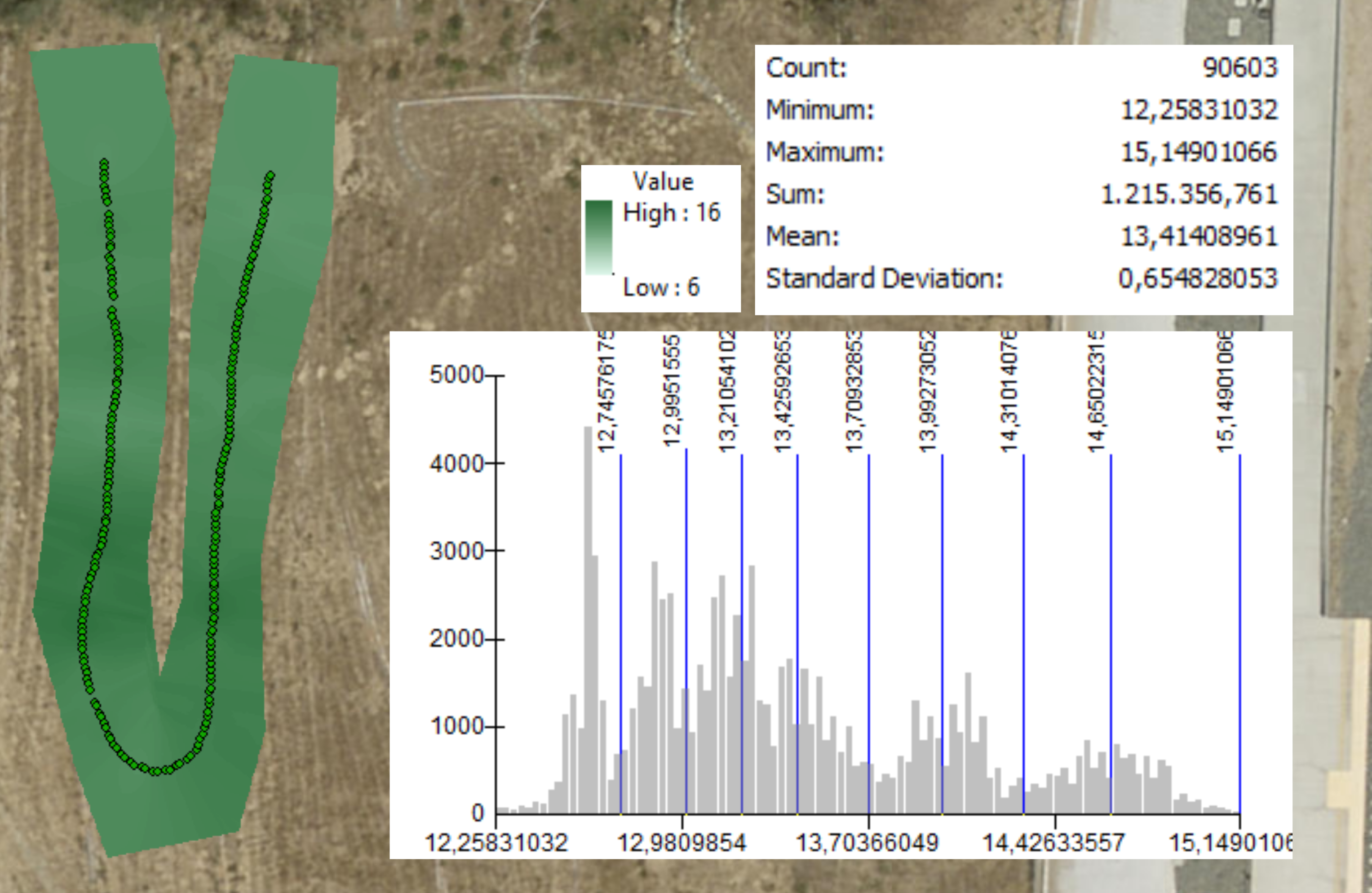}}
      \caption{(a) The soil conductivity curves of the U-shaped trajectory case in the bare field. Fitted graphs of both measurement curves correspond to an 8th grade least squares fit. (b)-(c) Soil ECa maps corresponding to manually-collected (handheld) and robotized-collected data, respectively, computed by applying kriging interpolation through exponential semivariogram. Each panel also contains the value-based color scale, the map statistics, and the histogram of the conductivity values. \label{fig:barevalidation40}}
      \vspace{-6pt}
\end{figure}

Obtained results visualized based on aggregated soil conductivity plots and the corresponding ECa maps for the cases of $d_b=70\;cm$ (straight line trajectory) and $d_b=40\;cm$ (U-shaped trajectory) are depicted in Fig.~\ref{fig:barevalidation70} and Fig.~\ref{fig:barevalidation40}, respectively.
Foremost, results validate that the shorter probe distance placement demonstrates a higher difference on average sensor readings compared to the longer distance setting against the manual baseline data (case $d_b=40\;cm$, robotized: $\{\mu=13.32, \sigma=0.99\}\;mS/m$ and manual: $\{\mu=7.55, \sigma=0.92\}\;mS/m$; case $d_b=70\;cm$, robotized: $\{\mu=11.40, \sigma=1.83\}\;mS/m$ and manual: $\{\mu=9.34, \sigma=1.59\}\;mS/m$). Despite this increased difference, however, the standard deviations are close between robotized and manual measurements in both cases. This indicates that there is consistency among the two types of measurements, and that the observed offsets in robotized measurements compared to their respective manual baselines can be in fact treated as constant offsets. Further evidence in support of the constant offset presence can be obtained by the Pearson Correlation Coefficient (PCC). In both cases (straight line and U-shape), the robotized ECa measurements showcase high linearity with their manual counterparts. Specifically, by removing outlier measurements that lie out of the $\pm 2\sigma$ measurement distribution, the straight-line trajectory attains a PCC of 0.95 whereas the U-shaped trajectory reaches a PCC of 0.88 (despite significant electromagnetic interference as demonstrated in static tests discussed in Section~\ref{seq:sensorinterference}), compared with manual measurements that were conducted on exactly the same paths. These findings can be visually corroborated by the obtained ECa maps in panels (b) and (c) of Fig.~\ref{fig:barevalidation70} and Fig.~\ref{fig:barevalidation40}, with the main difference being that the color differential in the shorter probe placement case of $d_b=40\;cm$ (Fig.~\ref{fig:barevalidation40}) being more pronounced due to the larger constant offset in measurements compared to the longer probe placement case of $d_b=70\;cm$ (Fig.~\ref{fig:barevalidation70}).

In all, these preliminary feasibility testing validates we can perform trustworthy continuous ECa measurements with the developed robotic setup even at the two boundary configurations. Results verify the system's high linearity with respect to manually-collected (handheld) data, and that the constant additive offset on the overall ECa measurements caused by the robot's electromagnetic interference does not significantly affect the linearity of obtained measurements in the end. Putting everything together, we conclude that the configuration $\{d_b=60\;cm, d_h=6\;cm\}$ can serve as an appropriate tradeoff that combines lower robot-to-sensor signal interference and higher maneuverability, and it is thus selected as the optimal configuration to conduct field-scale experiments. These are discussed next.

\section{Field-scale Experiments}\label{sec:evaluation}

The analysis conducted so far has helped determine an optimal sensor setup ($\{d_b=60\;cm, d_h=6\;cm\}$) for the robot considered in this study that balances between electromagnetic interference caused by the robot and its electronic components and robot traversability capacity of uneven terrain. The analysis has also helped validate the preliminary feasibility of collecting continuous soil ECa measurements over small bare-field areas, with obtained results being trustworthy and consistent to manually-collected baselines. We now turn our attention to robot-assisted continuous soil ECa measurements over larger fields.

\subsubsection*{\textbf{Experimental Procedure}} 
\begin{figure}[!t] 
    \centering
    \subfloat[\label{jackalatagops}]{%
        \includegraphics[width=0.37\linewidth]{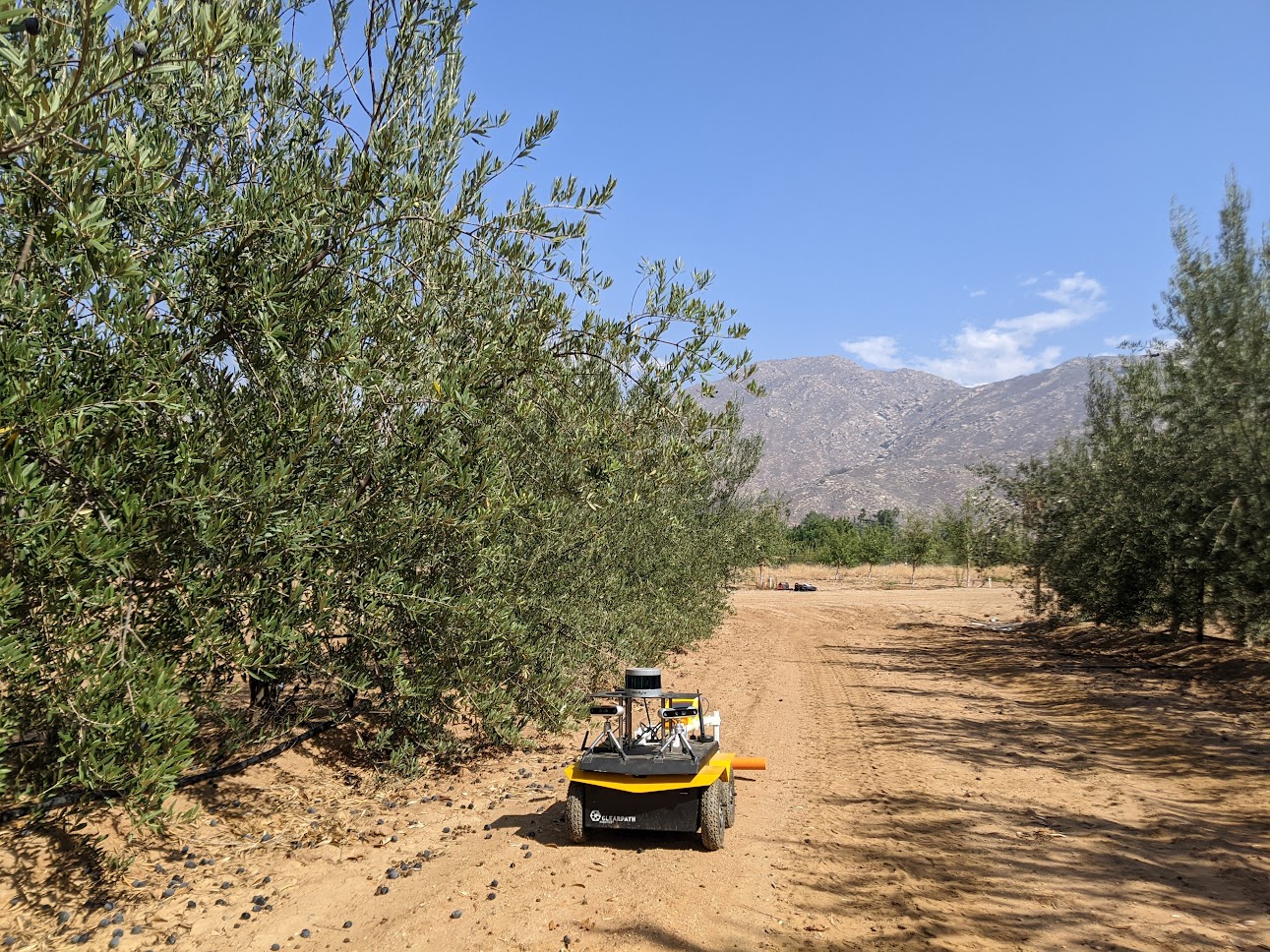}}
    \hfill
    \subfloat[\label{agops_me_doing}]{%
       \includegraphics[trim={0cm, 0cm, 0cm, 5.7cm},clip,width=0.25\linewidth]{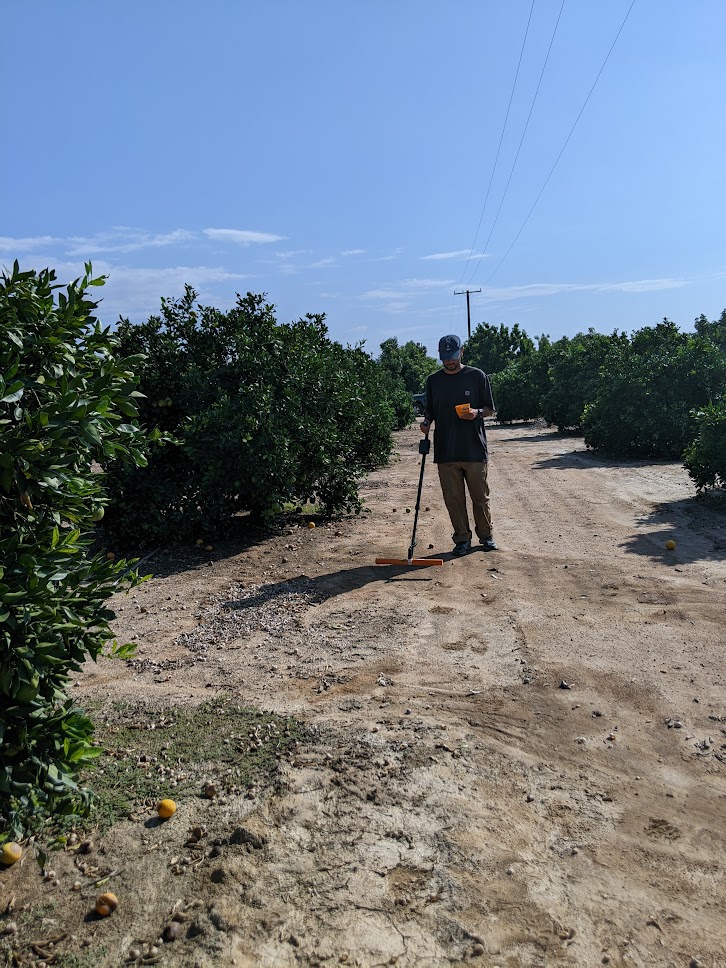}}
    \hfill
    \subfloat[\label{agops_jackal}]{%
        \includegraphics[width=0.37\linewidth]{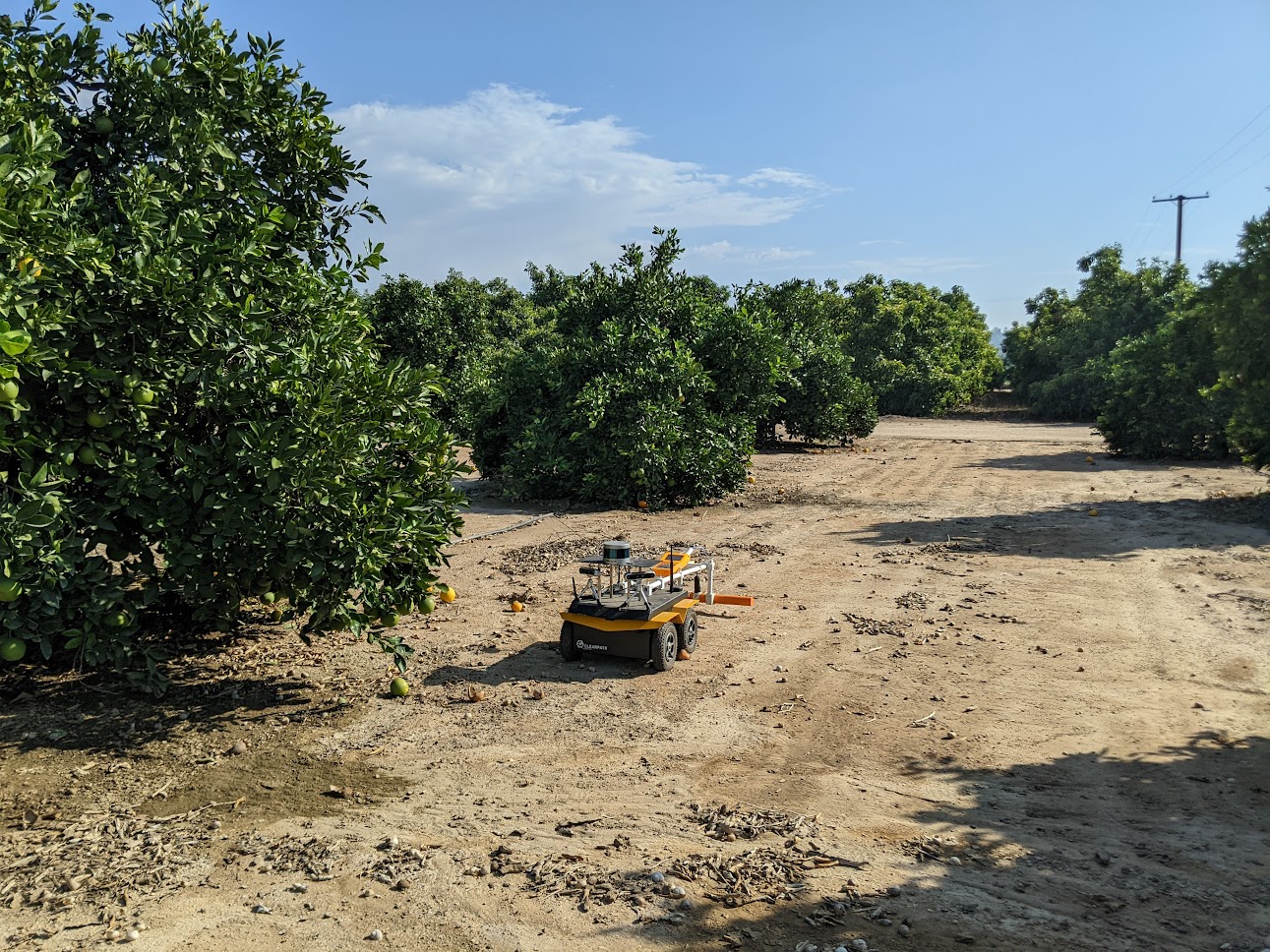}}
  \caption{Field-scale data collections instances: (a) robotized in the olive tree grove and (b)/(c) manual/robotized in the orange tree grove.\label{fig:systemevaluationsetup}}
\end{figure}

We performed continuous soil ECa measurements in two distinctive fields, an olive tree grove (not irrigated recently with respect to data collections) and a citrus tree grove (irrigated prior to data collections). The arid canopy of olive trees is located close to the USDA-ARS U.S. Salinity Laboratory at the University of California, Riverside (UCR) (33$^\circ$58'21.936'' N, -117$^\circ$19'13.5732'' E), whereas the freshly irrigated citrus orchard is located within the UCR \modd{Agricultural Experimental Station} fields (\add{AES; }33$^\circ$57'52.0272'' N, -117$^\circ$20'13.7184'' E). It is worth noting that the latter case in fact contained various soil conditions (highly irrigated/wet parts and arid parts), and hence helped evaluate 1) the proposed robot's performance in the same survey as terrain conditions vary, and 2) if the robot-assisted soil conductivity curves and soil ECa maps match those corresponding to manual data collections. 
Experimental setups and snapshots of the two fields are depicted in Fig.~\ref{fig:systemevaluationsetup}. For the olive tree grove, we considered an area of two full tree rows covered following a U-shaped trajectory, whereas, for the citrus tree grove, we considered an area of roughly \modd{three tree} rows covered following an S-shaped trajectory. Manual data collections with the handheld sensor (three in each field) were also performed to serve as the baseline. In total, we conducted 12 experimental trials for overall system field-scale evaluation in the olive and citrus tree groves (six robotized and six manual). Obtained soil ECa measurements were georeferenced via RTK-GNSS in the robotized measurement and the sensor's embedded GPS in the manual measurements. Collected soil ECa data were used to generate a custom raster of the surveyed area; then kriging interpolation through exponential semivariogram helped obtain the ECa map which was embedded onto satellite imagery via the ESRI ArcGIS 10.8.2 software.

\subsubsection*{\textbf{Key Findings}} 
\begin{figure}[!t]
    \centering
      \subfloat[\label{salinity_handheld}]{%
    \includegraphics[trim={0.85cm, 0cm, 1.5cm, 1cm},clip,width=0.32\linewidth]{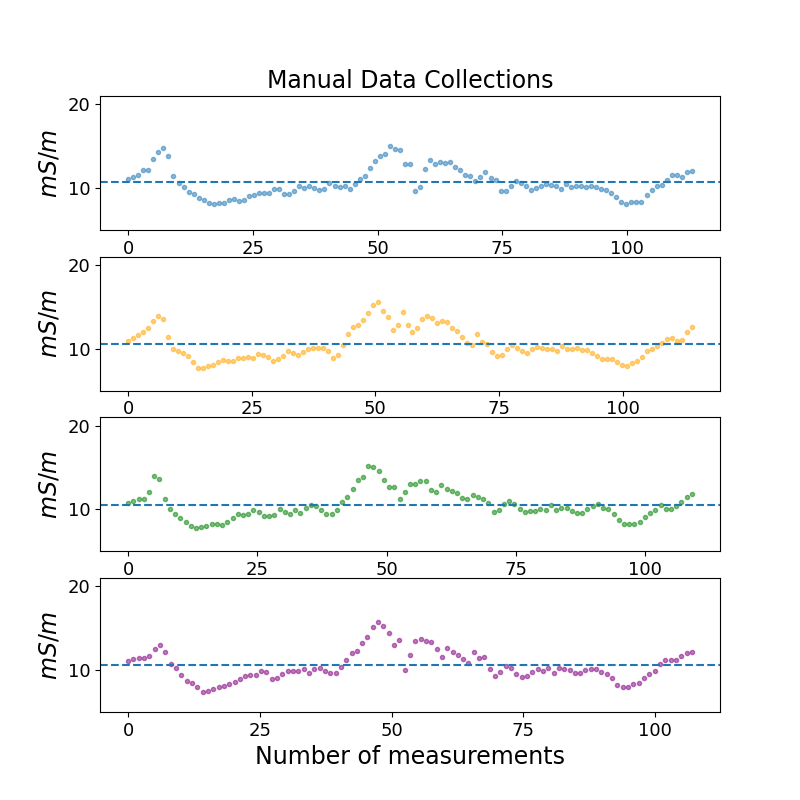}}
           \hfill
      \subfloat[\label{salinity_robot}]{%
    \includegraphics[trim={0.85cm, 0cm, 1.5cm, 1cm},clip,width=0.32\linewidth]{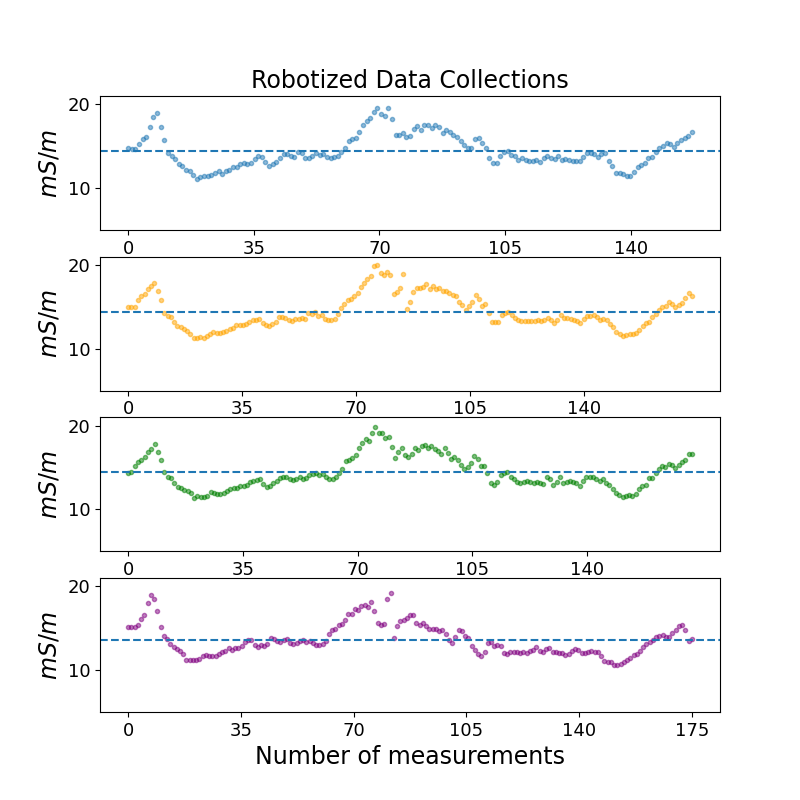}}
    \hfill
  \subfloat[\label{salinity_mean}]{%
       \includegraphics[trim={1.35cm, 0cm, 2.5cm, 1.5cm},clip,width=0.355\linewidth]{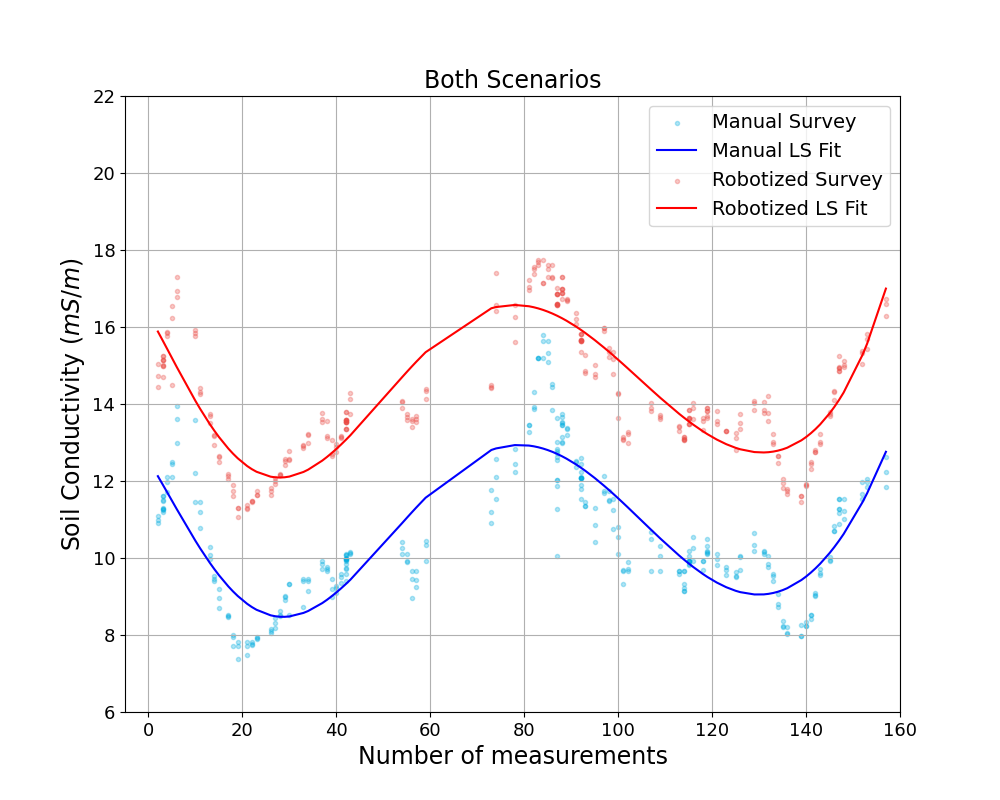}}
  \caption{Olive tree grove case. (a) and (b) Graphs of the raw conductivity measurements by using directly the sensor and via the robot, respectively. (c) Soil conductivity plots of both handheld and robot cases. Fitted plot of both measurement curves is an 8th grade least squares fit.\label{fig:olivePlots}}
  \vspace{-10pt}
\end{figure}

Results from field-scale experiments for olive and citrus tree groves are shown in Fig.~\ref{fig:olivePlots}--\ref{fig:citrusECa}. Collected raw data from each considered case, aggregated soil conductivity data comparing robotized to manual baselines, as well as the corresponding ECa maps are presented. 
It can be readily verified from the graphs that robot-assisted continuous ECa measurements approximate very well the manually-collected baselines in both cases.

\begin{figure}[!t]
    \centering
    \subfloat[\label{salinity_arcgis_handheld}]{%
        \includegraphics[width=0.49\linewidth]{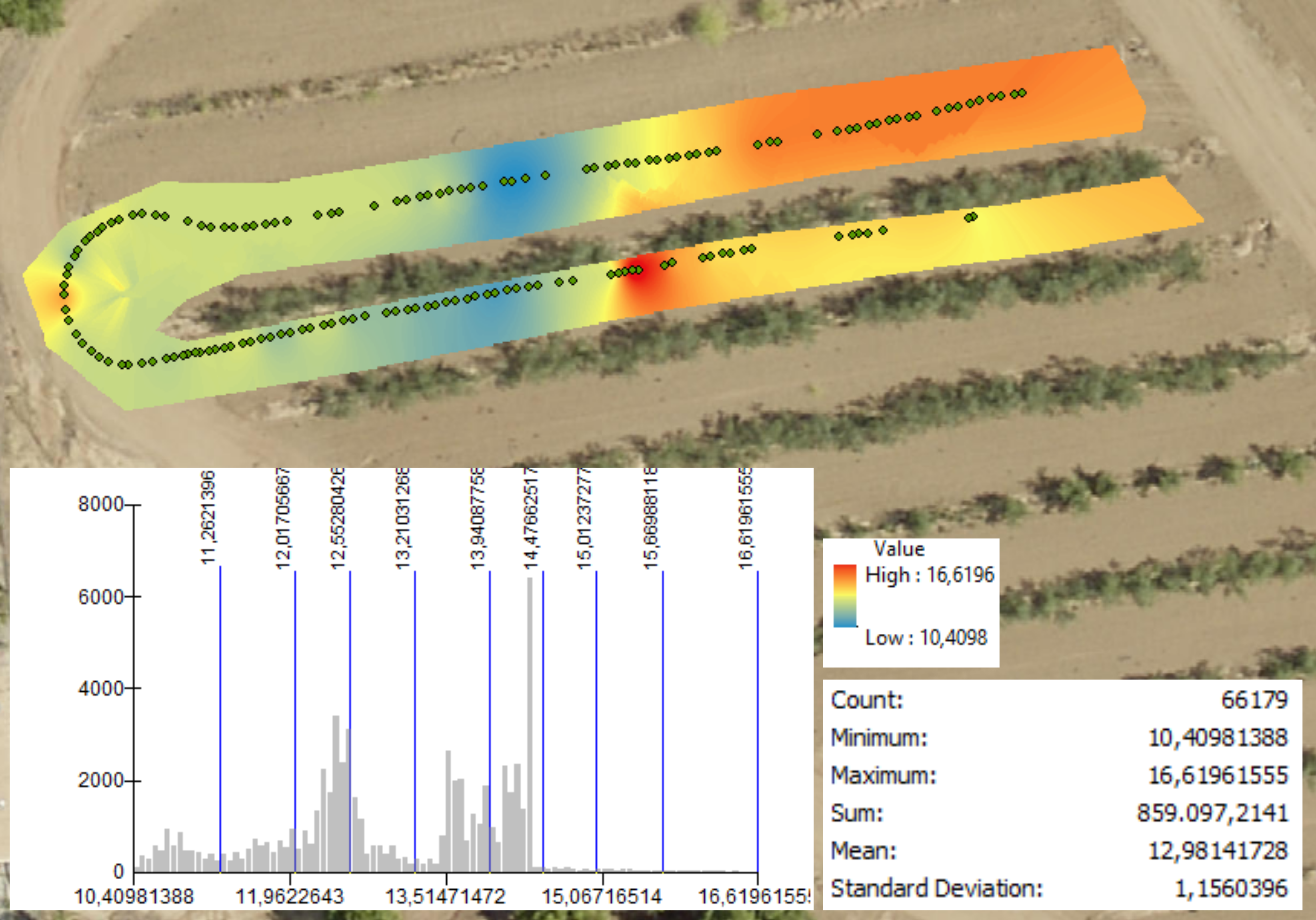}}
    \hspace{0.2cm}
\subfloat[\label{salinity_arcgis_jackal}]{%
    \includegraphics[width=0.49\linewidth]{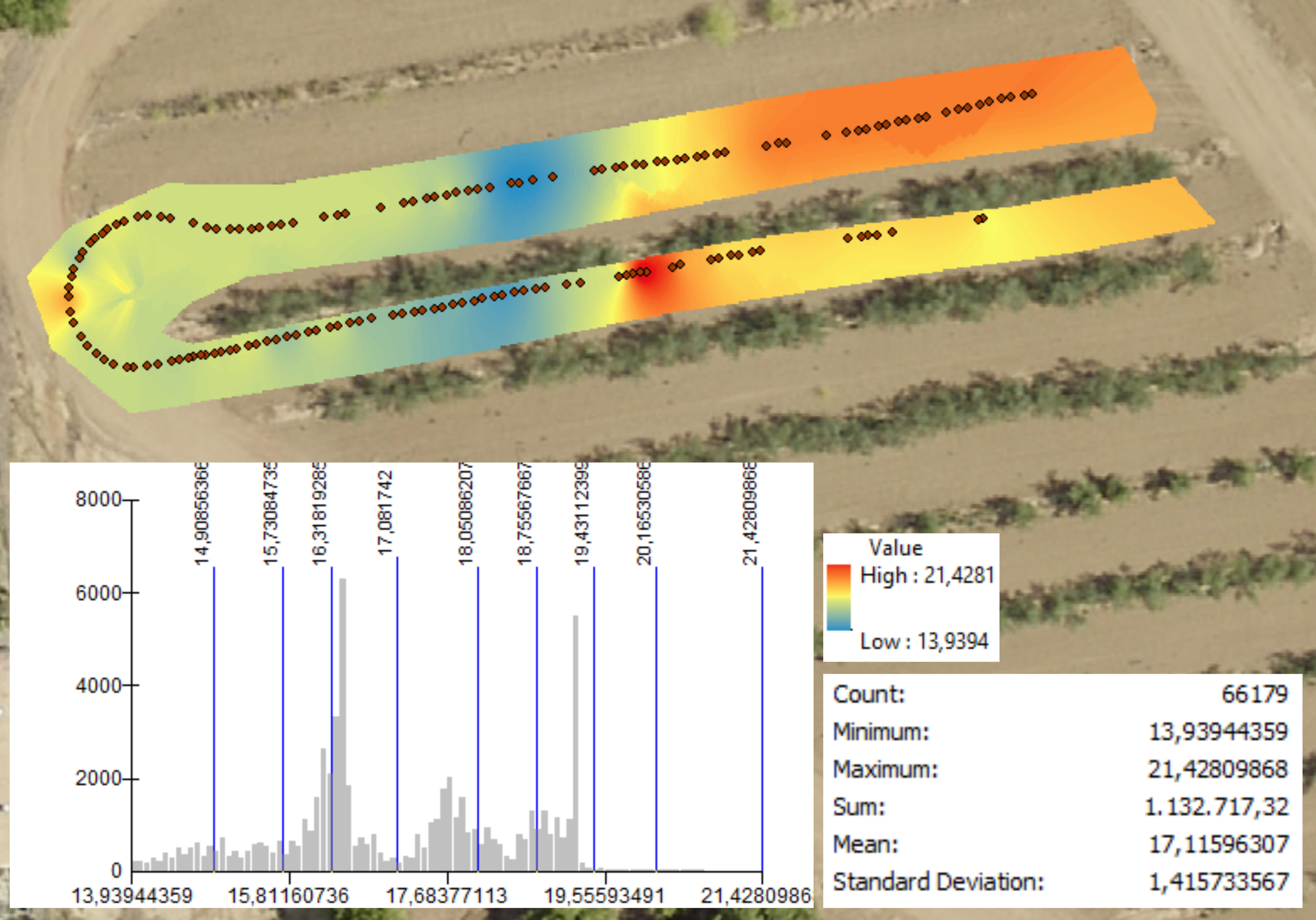}}
  \caption{Obtained soil ECa maps in the olive tree grove for (a) the manual and (b) robotized cases. Maps have been created by applying kriging interpolation through exponential semivariogram. Each map also depicts the value-based color scale, the map statistics, and the histogram of the conductivity values.\label{fig:oliveECa}}
  \vspace{-10pt}
\end{figure}

For the case of the olive tree grove (arid field) raw measurement plots (Fig.~\ref{salinity_handheld} and Fig.~\ref{salinity_robot}) match either very well. The visual observation is corroborated via the mean measurement plots (Fig.~\ref{salinity_mean}), where PCC reaches a score of 0.97. The mean conductivity value in the robotized case is $14.23\;mS/m$, compared to the $10.56\;mS/m$ mean value of the manual case, which represents a fixed increase which can be seen by the curvature of the polynomial fits. Additionally, the standard deviations are closely matching robotized: $1.65\;mS/m$; manual: $1.75\;mS/m$) which demonstrates consistency across both soil ECa measurement means. Output ECa maps (Fig.~\ref{fig:oliveECa}) are also very similar, with a PCC value of 0.97 in a pixel-wise correlation. It is notably clear that the robotized results are close to the manual ones, in spite of the constant positive offset on the level of the measured soil apparent electrical conductivity.

Similar observations can be readily made for the case of the citrus tree grove. Recall that this field was recently irrigated prior to data collections, hence there was more terrain variability; this ranged from muddy soil to normally irrigated areas and more dry parts of the field areas. From a robot operation standpoint, the proposed system demonstrates robust behavior even in this diverse and more demanding survey case, and it is noteworthy that the robot's traversability in the irrigated turf was efficient even when navigating over mud. Raw data graphs (Fig.~\ref{agops_handheld} and Fig.~\ref{agops_robot}) demonstrate several notable peaks in soil conductivity measurements that were caused when traversability in the muddy terrain and the well-irrigated soil. These peaks were captured in both manual and robot-assisted surveys. The graphs shown in Fig.~\ref{agops_means} have a PCC of 0.90, which demonstrates the proposed robot's robustness in even muddy and quite diverse soil-ECa-level fields. The polynomial fits of both plots (Fig.~\ref{agops_means}) present similar curvature with an additive offset increase in the robot's case, caused by the electromagnetic interference, whereas the pixel-wise correlation of the output ECa maps reaches a value of 0.96 (Fig.~\ref{fig:citrusECa}). Visual inspection of the obtained soil ECa maps also supports the aforementioned findings. 

%% file: conclusion.tex
\section{Conclusion}\label{sec:conclusion}

In this study, we presented a robotized means to perform precise and continuous soil apparent electrical conductivity (ECa) measurement surveys. Our proposed solution involves a ground mobile robot equipped with a customized and adjustable platform to hold an Electromagnetic Induction (EMI) sensor. \add{The optimal placement of the EMI sensor is determined by satisfying two competing objectives; the minimization of static electromagnetic interference in measurements and the facilitation of robot traversability in the field.} Extensive experimental evaluation across static calibration tests and over different types of fields \add{concludes to the optimal EMI configuration setup to be used for large-field testing}. \add{Throughout a series of real field experiments, our study} demonstrates that the obtained robot-assisted soil conductivity measurements present high linearity compared to the ground truth (data collected manually by a handheld EMI sensor) by scoring more than $90\%$ in Pearson correlation coefficient in both plot measurements and estimated \modd{ECa} maps generated by kriging interpolation. The proposed platform can deliver high-linearity scores in real survey scenarios, at an olive and citrus grove under different irrigation levels, and serve as a robust tool for large-scale ECa mapping in the field, with the potential development of a fully-autonomous behavior. Future work will focus on integration within a task and motion framework for informative proximal sampling, as well as integration with physical sampling means to perform multiple tasks simultaneously.

\begin{figure}[!t]
    \centering
      \subfloat[\label{agops_handheld}]{%
    \includegraphics[trim={0.2cm, 0cm, 1.5cm, 0.5cm},clip,width=0.32\linewidth]{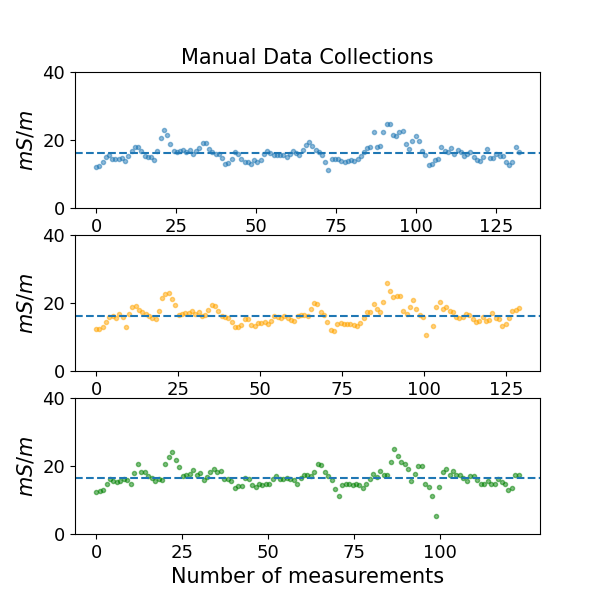}}
           \hfill
      \subfloat[\label{agops_robot}]{%
    \includegraphics[trim={0.2cm, 0cm, 1.5cm, 0.5cm},clip,width=0.32\linewidth]{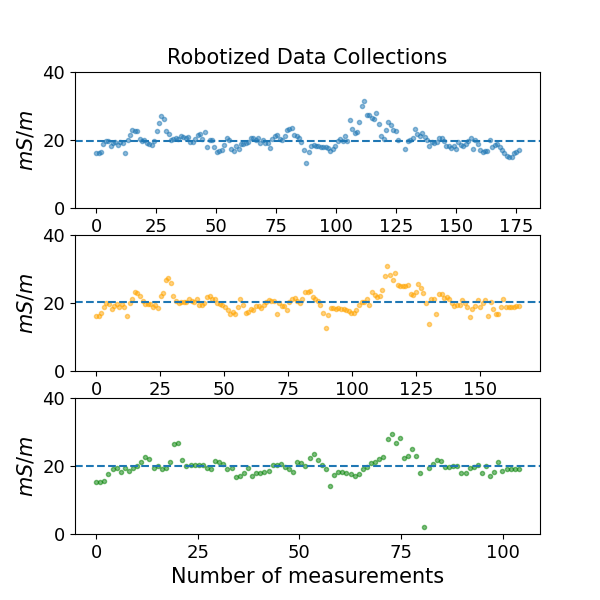}}
    \hfill
  \subfloat[\label{agops_means}]{%
       \includegraphics[trim={1cm, 0cm, 2.5cm, 1cm},clip,width=0.355\linewidth]{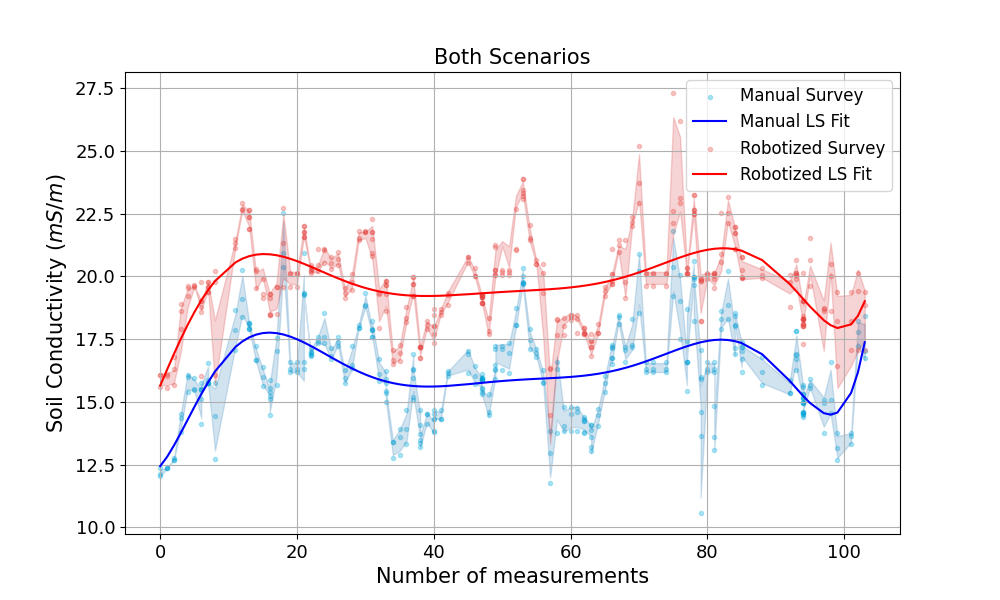}}
\caption{Citrus tree grove case. (a) and (b) Graphs of the raw conductivity measurements by using directly the sensor and via the robot, respectively. (c) Soil conductivity plots of both handheld and robot cases. \add{The color-shaded areas indicate the filled areas of soil conductivity values of the corresponding manual and robotized measurements.} Fitted plot of both measurement curves is an 8th grade least squares fit. \label{fig:citrusPlots}}
\vspace{0pt}
\end{figure}

\begin{figure}[!t]
    \centering
  \subfloat[\label{handheld_arcgis_agops}]{%
       \includegraphics[width=0.49\linewidth]{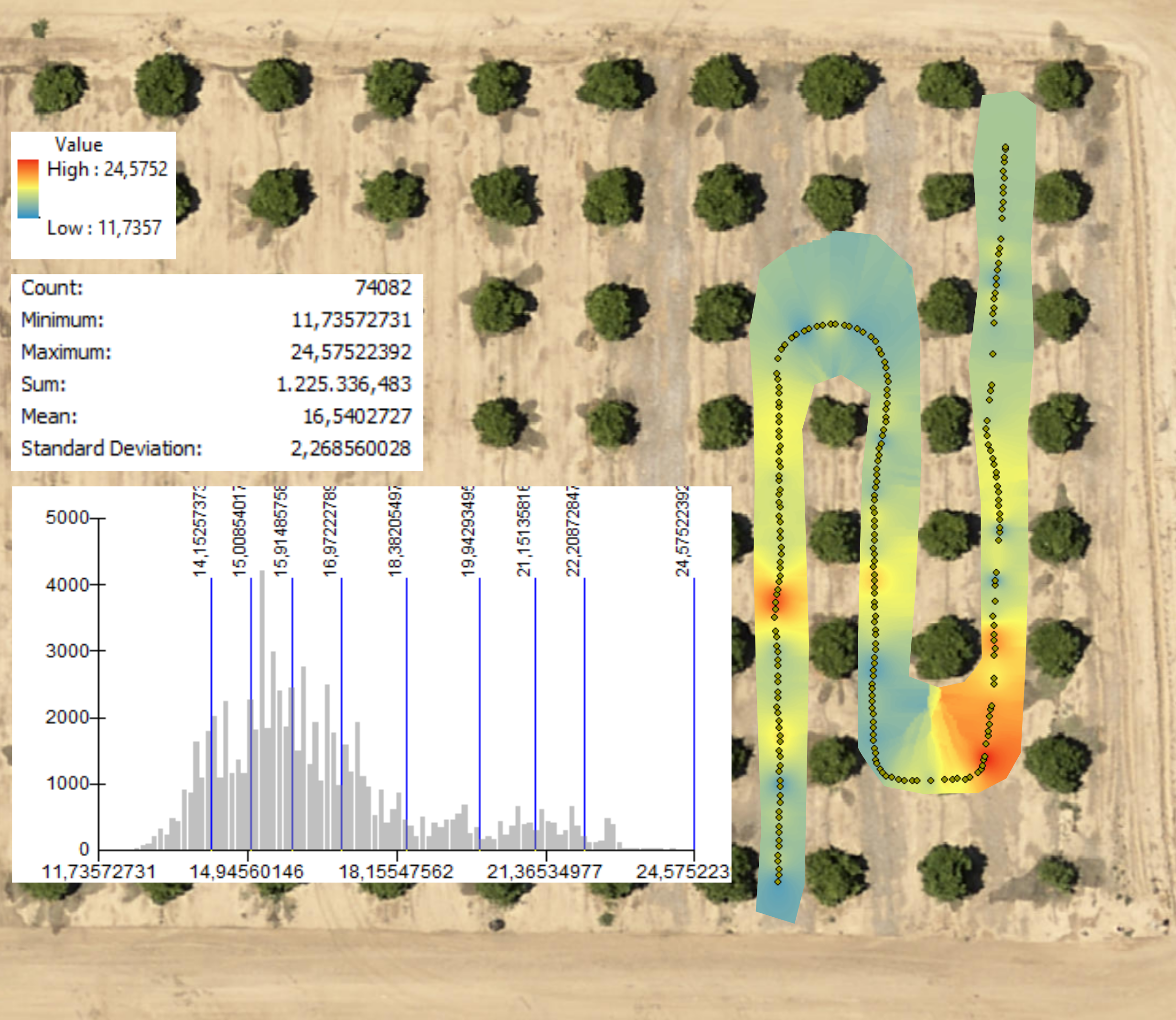}}
           \hspace{0.2cm}
      \subfloat[\label{jackal_arcgis_agops}]{%
    \includegraphics[width=0.49\linewidth]{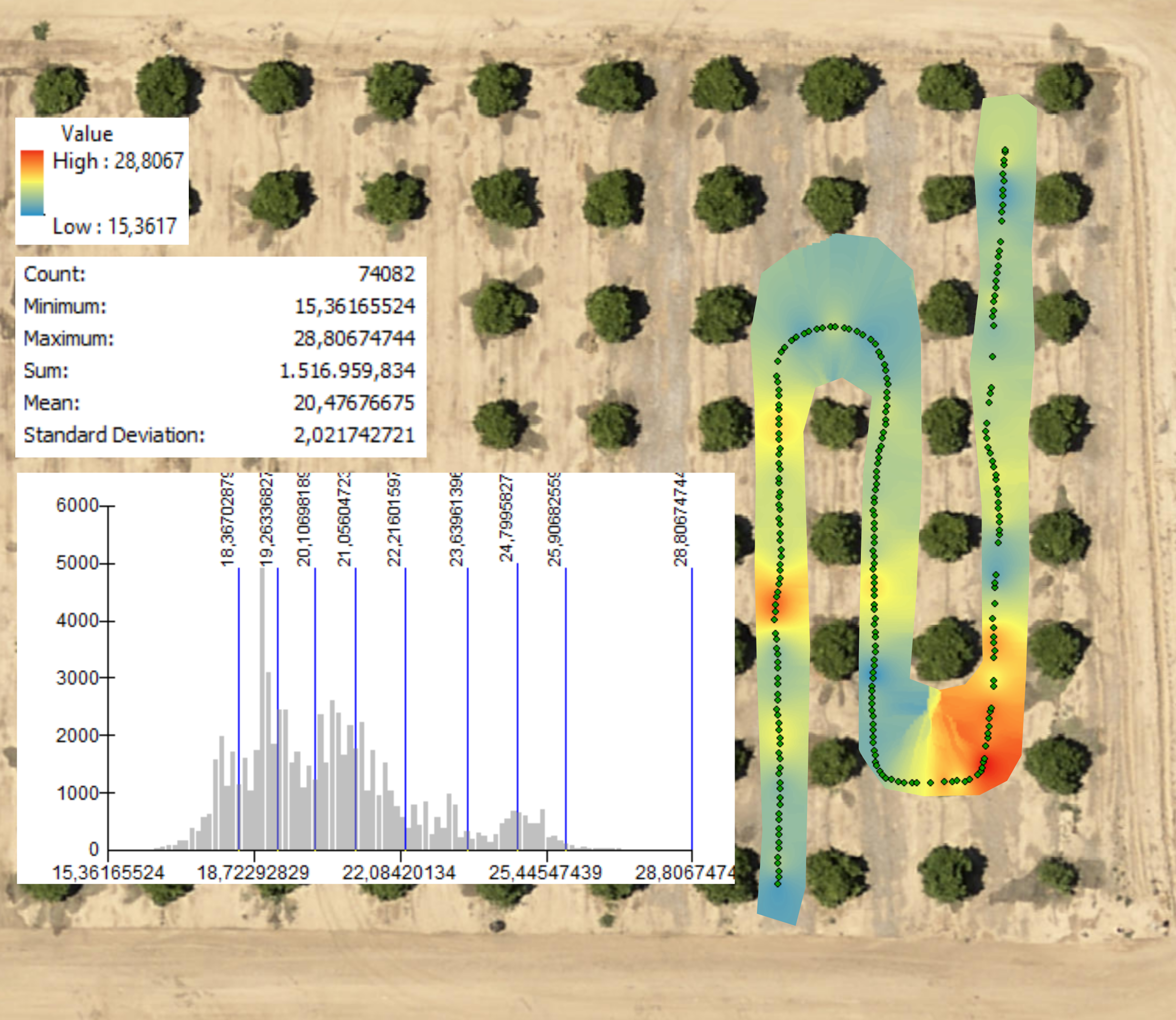}}
  \caption{Obtained soil ECa maps in the citrus tree grove for (a) the manual and (b) robotized cases. Maps have been created by applying kriging interpolation through exponential semivariogram. Each map also depicts the value-based color scale, the map statistics, and the histogram of the conductivity values.\label{fig:citrusECa}}
  \vspace{-10pt}
\end{figure}